\documentclass[10.5pt]{article}

\PassOptionsToPackage{numbers, compress}{natbib}
\usepackage[utf8]{inputenc}
\usepackage{amsmath}
\usepackage{amsfonts}
\usepackage{mathrsfs}
\usepackage{amssymb}
\usepackage{amsthm}

\usepackage{libertine}
\usepackage{wrapfig}
\usepackage{graphicx}
\usepackage{multirow}
\usepackage{epstopdf}
\usepackage{multicol}
\usepackage{subfigure}
\usepackage{booktabs}

\usepackage{algorithmic}
\usepackage{algorithm}
\usepackage[algo2e,linesnumbered]{algorithm2e}
\usepackage{stackengine}
\usepackage{graphicx}
\usepackage{multirow}
\usepackage{epstopdf}
\usepackage{multicol}
\usepackage{caption}
\usepackage{subfigure}
\usepackage{enumitem}
\usepackage{amsmath}
\usepackage{pifont}
\usepackage{natbib}
\usepackage{bm}
\usepackage{bbm}
\usepackage{makecell}
\usepackage{mathtools}
\usepackage[table]{xcolor}
\definecolor{mydarkred}{rgb}{0.6,0,0}
\definecolor{mydarkgreen}{rgb}{0,0.6,0}
\usepackage[colorlinks,
linkcolor=mydarkred,
citecolor=mydarkgreen]{hyperref}
\usepackage{url}
\usepackage{bm}
\usepackage{pifont}
\usepackage{stfloats}
\usepackage[T1]{fontenc}
\usepackage{arydshln}
\usepackage{colortbl}
\usepackage{balance}

\newtheorem{proposition}{Proposition}
\newcommand{\argmin}{\operatornamewithlimits{argmin}}

\newcolumntype{L}[1]{>{\raggedright\let\newline\\\arraybackslash\hspace{0pt}}m{#1}}
\newcolumntype{Y}{>{\centering\arraybackslash}X}
\newcolumntype{s}{>{\hsize=.3\hsize}Y}
\newcolumntype{t}{>{\hsize=1.5\hsize}X}
\newcolumntype{u}{>{\hsize=0.8\hsize}Y}
\usepackage{makecell}
\usepackage{mathtools}
\usepackage[utf8]{inputenc}
\usepackage{stfloats}

\title{Pluralistic Image Completion\\ with Probabilistic Mixture-of-Experts}

\author{
  Xiaobo Xia$^{1}\thanks{Equal contributions.}$,
  Wenhao Yang$^{2}\footnotemark[1]$,
  Jie Ren$^{3}$,\\
  Yewen Li$^4$,
  Yibing Zhan$^5$,
  Bo Han$^6$,
  Tongliang Liu$^1$\\
  \small{$^1$The University of Sydney;}
  \small{$^2$Nanjing University;}
  \small{$^3$The University of Edinburgh;}\\
  \small{$^4$Nanyang Technological University;}
  \small{$^5$JD Explore Academy;}
  \small{$^6$Hong Kong Baptist University}
}
\date{}
\begin{document}

\maketitle

\begin{abstract}
Pluralistic image completion focuses on generating both \textit{visually realistic} and \textit{diverse} results for image completion. Prior methods enjoy the empirical successes of this task. However, their used constraints for pluralistic image completion are argued to be \textit{not well interpretable} and \textit{unsatisfactory} from two aspects. First, the constraints for visual reality can be \textit{weakly correlated} to the objective of image completion or even \textit{redundant}. Second, the constraints for diversity are designed to be \textit{task-agnostic}, which causes the constraints to not work well. In this paper, to address the issues, we propose an end-to-end probabilistic method. Specifically, we introduce a unified \textit{probabilistic graph model} that represents the complex interactions in image completion. The entire procedure of image completion is then mathematically divided into several sub-procedures, which helps efficient enforcement of constraints. The sub-procedure directly related to pluralistic results is identified, where the interaction is established by a \textit{Gaussian mixture model} (GMM). The inherent parameters of GMM are \textit{task-related}, which are \textit{optimized adaptively} during training, while the number of its primitives can control the diversity of results conveniently. We formally establish the effectiveness of our method and demonstrate it with comprehensive experiments.
\end{abstract}

\section{Introduction}

Pluralistic image completion refers to the task of filling in the missing region of an incomplete image, so as to produce \textit{visually realistic} and \textit{diverse} image completion solutions~\cite{zheng2019pluralistic,zheng2021pluralistic,zheng2021tfill}. Different from single image completion \cite{ren2015shepard,grigorev2019coordinate,zeng2021cr,yi2020contextual} that learns a \textit{deterministic} mapping from an incomplete image to a complete image, and produces a \textit{unique} result, pluralistic image completion can generate \textit{various} results with visually realistic contents. Pluralistic image completion follows the fact that image completion is a \textit{highly subjective} process and benefits a series of applications such as photo restoration \cite{liu2021pd}, object removal \cite{han2019finet}, and transmission error concealment \cite{peng2021generating}.

Pluralistic image completion is complicated and challenging. The state-of-the-art methods encode images to latent features. To constrain the visual reality of results, \textit{heuristic constraints} are introduced, \textit{e.g.}, the perceptual loss \cite{liu2021pd} and attention module \cite{zheng2019pluralistic,wan2021high}. To constrain the diversity of results, the distribution of latent features is assumed to be a \textit{unimodal} Gaussian distribution with \textit{predefined parameters} \cite{zheng2019pluralistic}. Image completion is performed by \textit{sampling} from the unimodal Gaussian distribution. 

Although the above paradigm enjoys empirical successes of pluralistic image completion, it is \textit{not well interpretable} and \textit{unsatisfactory} in practice. We detail the issues from two aspects. First, the added constraints for the visual reality of results are based on \textit{general purposes}. These constraints may work well in some tasks. Unfortunately, there is no clear understanding of what role they play in pluralistic image completion. Inappropriate constraints are likely to have side effects. It is not easy to finish the determination of added constraints as our desideratum, especially in complex image completion tasks. Second, the diversity of image completion results is hard to be constrained in reason. For a specific task, the parameters of the unimodal Gaussian distribution are directly related to the diversity. Nevertheless, these parameters are designed to be \textit{prior knowledge} and \textit{task-agnostic}, which cause \textit{unreasonable pluralistic results}.

\begin{wrapfigure}{r}{0.5\textwidth}
\begin{minipage}{0.5\textwidth}
\vspace{-17pt}
\begin{figure}[H]
    \centering
    \includegraphics[width=0.98\textwidth]{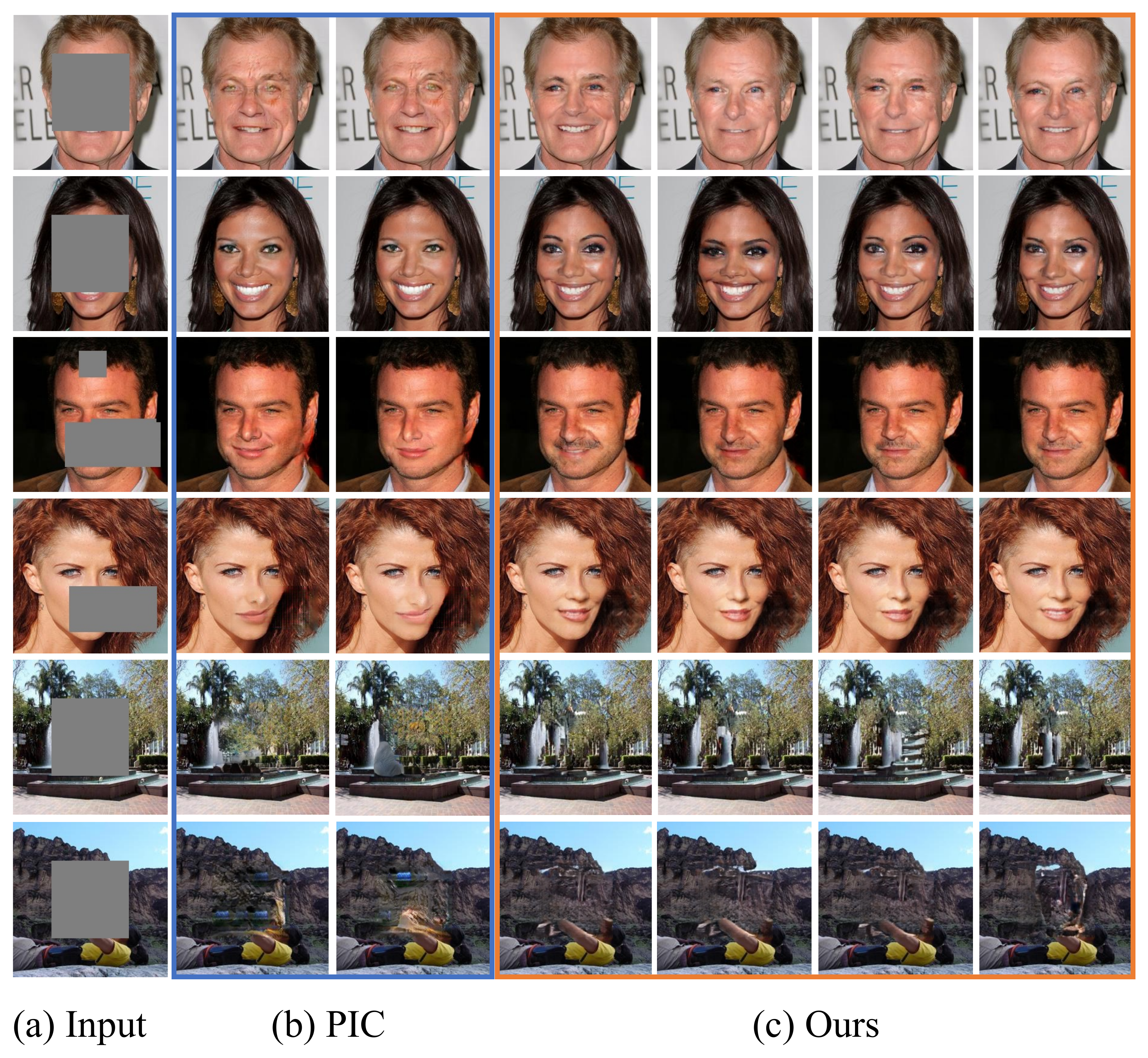}
    \vspace{-5pt}
    \caption{\textbf{Example completion results of the proposed method on the images of faces and natural sceneries with various masks} (missing regions are shown in gray). The baseline PIC refers to \cite{zheng2019pluralistic}. For each group, the masked input image is shown left, followed by diverse and plausible completion results from our method without any post-processing. (Zoom in to see the details.)}
    \label{fig:intro}
    \vspace{-5pt}
\end{figure}
\end{minipage}
\end{wrapfigure}

In this paper, to address the above issues, we present an end-to-end probabilistic method. The method is constructed with a unified \textit{probabilistic graph model} \cite{jordan2003introduction,koller2009probabilistic} and translates the problem of pluralistic image completion into a structured mathematical representation. Specifically, we suppose that pluralistic image completion can be represented with a directed acyclic graph that is designed reasonably. The nodes of the graph denote different kinds of images and corresponding latent features. The edges of the graph represent the interactions between nodes. Based on this directed acyclic graph, the entire procedure of pluralistic image completion is divided into several sub-procedures. Then, we perform different constraints for different sub-procedures as our desideratum.

In particular, the diversity of image completion results is implemented by modeling the interaction between the latent features of the missing region and incomplete image with a \textit{Gaussian mixture model} (GMM) \cite{reynolds2009gaussian,zong2018deep}. Compared with the unimodal Gaussian distribution, GMM has a larger capacity and is more competitive to meet the output diversity \cite{zhang2007probabilistic,ren2021probabilistic}. For our method, the inherent parameters of GMM are \textit{task-related} and \textit{optimized adaptively} during training. The different primitives of GMM represent the outputs with different patterns. Additionally, the number of primitives can be chosen artificially with the needs for diversity in real-world applications. 

Before delving into details, we clearly emphasize our
contributions as follows:
\begin{itemize}[leftmargin=*]
    \item To our best knowledge, this paper is the first one that uses a PGM for pluralistic image completion. Besides, the PGM is specially designed with justifications, rather than borrowing existing models. With the designed PGM, the added constraints for visual reality and diversity are explainable.
    \item We propose to use GMM to increase result diversity. One remarkable advantage is that the inherent parameters for diversity are task-related, leading to more reasonable results. 
    \item We conduct a series of experiments on benchmark datasets to support our claims. In both qualitative and quantitative comparisons with the state-of-the-art methods, our method achieves superior performance. The generated contents for image completion are both visually realistic and diverse, such as those shown in Figure \ref{fig:intro}. Codes will be released for the reproducibility of results. 
\end{itemize}

The rest of the paper is organized as follows. In Section \ref{sec:2}, we briefly review pluralistic image completion. In Section \ref{sec:3}, we discuss the proposed method step by step. Experimental results are provided in Section \ref{sec:4}. Finally, we conclude the paper in Section \ref{sec:5}. 
\newpage
\section{Preliminaries}\label{sec:2}
\begin{wrapfigure}{r}{0.5\textwidth}
\begin{minipage}{0.5\textwidth}
\vspace{-15pt}
\begin{figure}[H]
    \centering
    \includegraphics[width=0.85\textwidth]{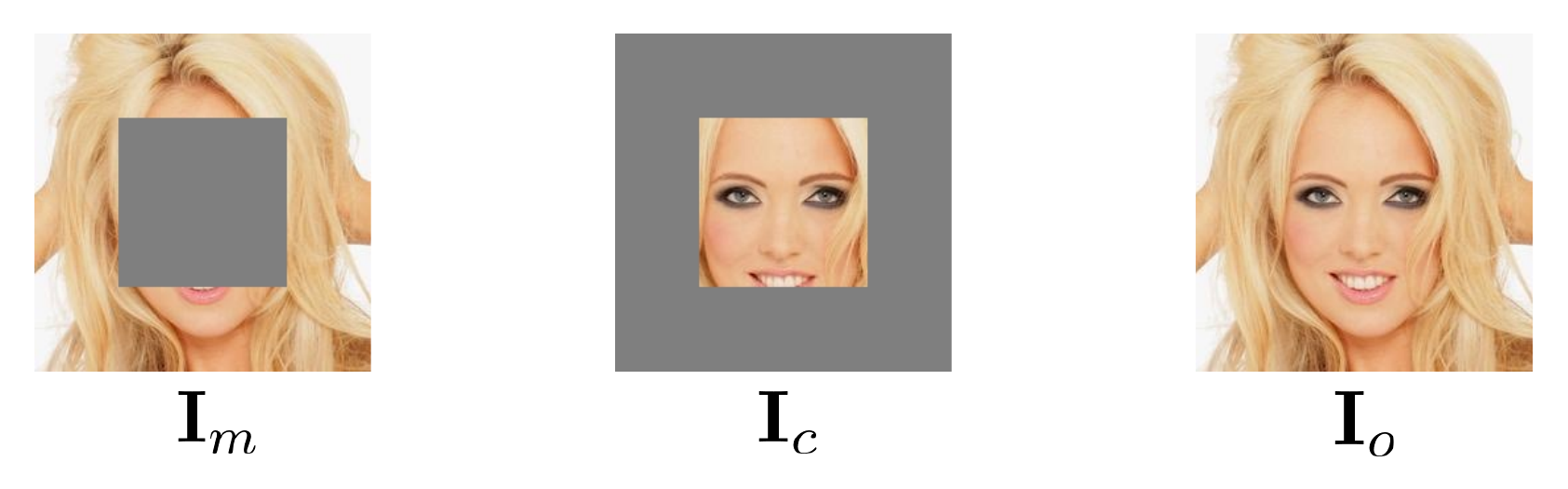}
    \caption{The illustrations for different kinds of images in image completion.}
    \vspace{-7pt}
    \label{fig:I}
\end{figure}
\end{minipage}
\end{wrapfigure}

\textbf{Problem Setting.} We first discuss image completion. For the task of image completion, suppose we have an image, originally $\mathbf{I}_o$, but degraded by some missing pixels to become a masked partial image $\mathbf{I}_m$. The image that comprises the original missing pixels is called a complement partial image, which can be denoted by $\mathbf{I}_c$. The goal of the image completion task is to reconstruct $\mathbf{I}_o$ by using $\mathbf{I}_m$. The illustrations for different kinds of images are provided in Figure~\ref{fig:I}. Traditional single image completion reconstructs the original image in a \textit{deterministic} fashion, which results in only a single solution \cite{grigorev2019coordinate,dong2020fashion}. 

Pluralistic image completion \cite{zheng2019pluralistic} is motivated by the fact that image completion is a highly subjective process. Also, completed images by different experts may agree on high-level semantics, but have substantially different details \cite{zheng2019pluralistic}. Therefore, in pluralistic image completion, we need to generate \textit{multiple} and \textit{diverse} plausible results when presented with a masked partial image. The problem of pluralistic image completion is challenging since we need to take care of the diversity and visual authenticity of results at the same time.

\textbf{Prior Work.} Traditional methods on single image completion, \textit{e.g.,} diffusion-based methods \cite{bertalmio2000image,ballester2001filling,levin2003learning} and patch-based methods \cite{barnes2009patchmatch,bertalmio2003simultaneous,criminisi2004region}, assume that image holes share similar content to visible regions. Therefore, they choose to directly propagate the contextual appearances or realign the background patches to complete the image holes. Benefiting from the strong power of deep neural networks, deep learning based single image completion methods achieve promising completion performance, \textit{e.g.,} employing shepard convolutional neural networks \cite{ren2015shepard}, restraining global and local consistency between image holes and visible regions \cite{iizuka2017globally}, performing deep feature rearrangement \cite{yan2018shift}, and learning a pyramid-context encoder network \cite{zeng2019learning}, \textit{etc}. In addition to these methods, multiple single image completion methods involve the generative adversarial networks \cite{goodfellow2014generative} to learn the semantic of images \cite{yang2019adaptive,yang2021objects}, \textit{e.g.,} \cite{li2017generative,yu2018generative,dolhansky2018eye,deng2018uv,zhang2018semantic,grigorev2019coordinate,dong2020fashion}. Although these single-solution methods achieve outstanding performance in predicting deterministic result for image holes, they cannot generate various semantically meaningful results.

Existing works for pluralistic image completion usually rely on generative models, \textit{e.g.}, CVAE \cite{walker2016uncertain,han2019finet,zheng2019pluralistic}, hierarchical VQ-VAE \cite{peng2021generating}, PD-GAN \cite{liu2021pd}, and BicycleGAN \cite{zhu2017multimodal}. Generative pluralistic image completion methods diversiform meaningful completion results with different kinds of constraints. However, such methods are not well interpretable as discussed. The issues largely limit their applications in the real world.

\section{Methodology}\label{sec:3}
This section presents a probabilistic method for pluralistic image completion. We first design a probabilistic graph
model for our task (Section \ref{sec:3.1}). Then, based on the probabilistic graph model, we show how to divide the entire procedure of pluralistic image completion into several sub-procedures (Section \ref{sec:3.2}). Afterward, we discuss how to use GMM to diversify image completion results (Section \ref{sec:3.3}). Finally, we add reconstruction and adversarial losses to strengthen image completion, and summarize all algorithm flows (Section \ref{sec:3.4}).  

\subsection{Probabilistic Graph Model Construction}\label{sec:3.1}
\begin{figure}[!t]
    \centering    
    \subfigure[\scriptsize Inference]{				
    \label{fig:pgm_a}							
    \includegraphics[width=0.24\textwidth]{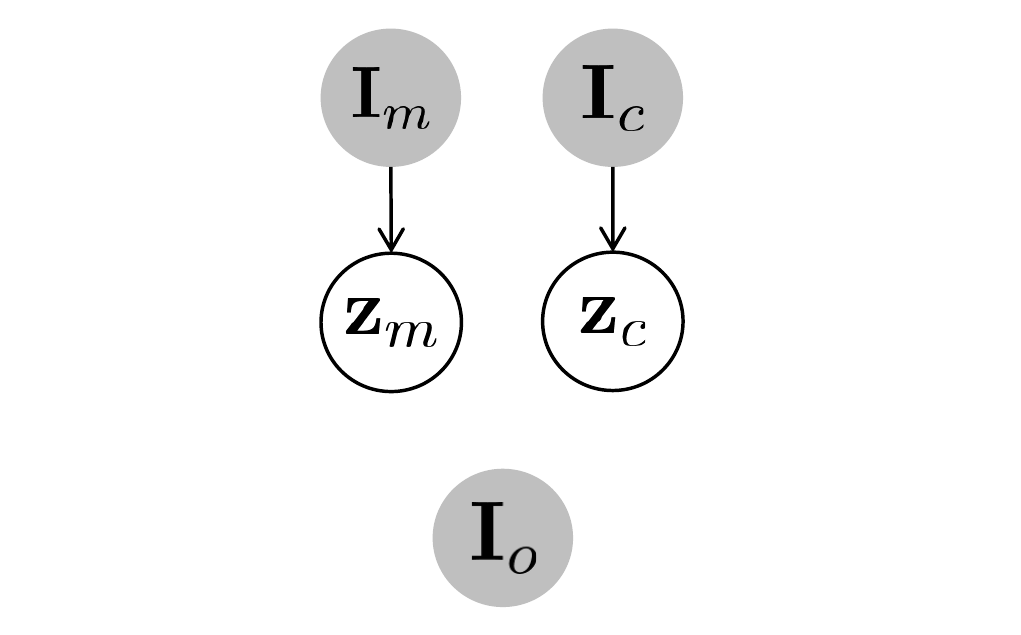}}\hspace{1pt}
    \subfigure[\scriptsize Dynamic Inference]{				
    \label{fig:pgm_b}						
    \includegraphics[width=0.24\textwidth]{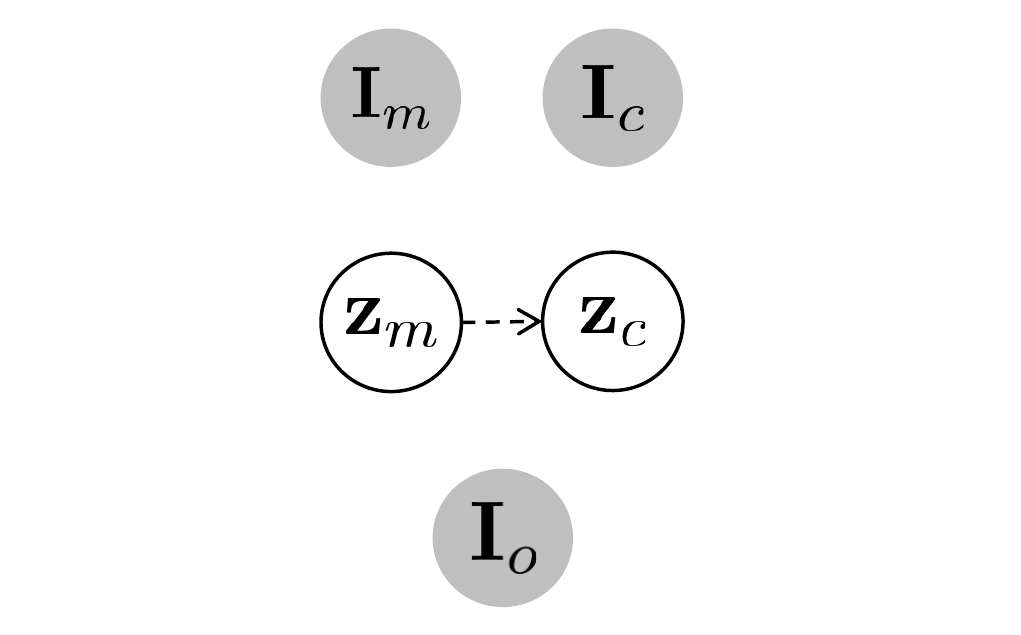}}\hspace{1pt}
    \subfigure[\scriptsize Generation]{				
    \label{fig:pgm_c}						
    \includegraphics[width=0.24\textwidth]{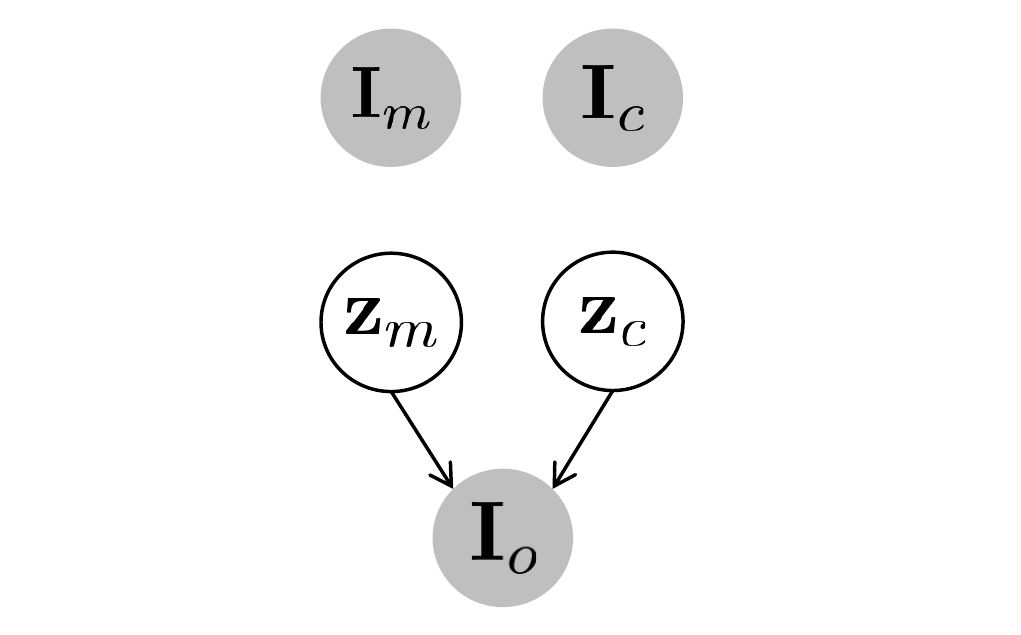}}\hspace{1pt}
    \subfigure[\scriptsize Overall]{				
    \label{fig:pgm_d}						
    \includegraphics[width=0.24\textwidth]{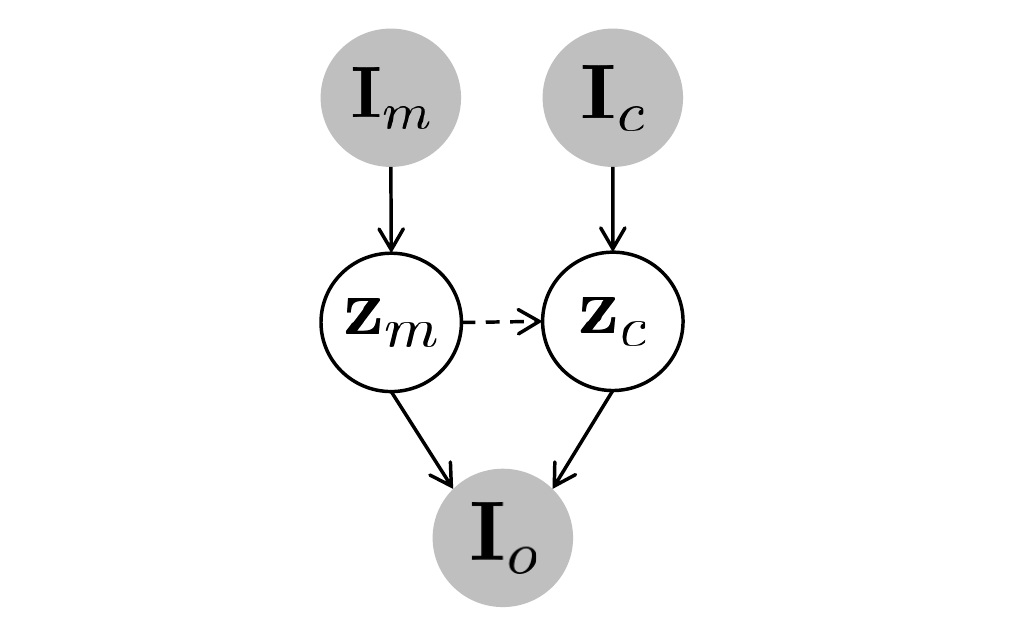}}
    \vspace{-5pt}
    \caption{The illustrations of the probabilistic graph model for the pluralistic image completion task.}		
    \label{fig:pgm}						
\end{figure}

Recall the notations in image completion, \textit{i.e.}, $\mathbf{I}_o$, $\mathbf{I}_m$, and $\mathbf{I}_c$, they correspond to the latent features $\mathbf{z}_o$, $\mathbf{z}_m$, and $\mathbf{z}_c$ respectively. We suppose a probabilistic graph model for the image completion task, which is shown in Figure \ref{fig:pgm}. The probabilistic graph model is designed reasonably. The graphical illustration of overall computational paths of the probabilistic graph model is presented in Figure~\ref{fig:pgm_d}. Specifically, the images $\mathbf{I}_m$ and $\mathbf{I}_c$ form $\mathbf{I}_o$. Their latent features $\mathbf{z}_m$ and $\mathbf{z}_c$ can be exploited to form $\mathbf{I}_o$. Given $\mathbf{z}_m$, we can surmise $\mathbf{z}_c$. We use the latent features here, since they are low-dimensional and informative~\cite{goodfellow2016deep,kingma2013auto}. We further details Figure \ref{fig:pgm} as follows.

\textbf{Inference.} We do not directly map $\mathbf{I}_m$ and $\mathbf{I}_c$ to $\mathbf{I}_o$, since they are too high-dimensional and have information redundancy. We exploit generative models. The processes $\mathbf{I}_m\rightarrow\mathbf{z}_m$ and $\mathbf{I}_c\rightarrow\mathbf{z}_c$ denote that we \textit{encode} images into corresponding latent features. During training, we learn the encoder $f$ to finish the inference. 

\textbf{Dynamic Inference.} Given $\mathbf{I}_m$ (\textit{resp.} $\mathbf{I}_c$), we can infer the latent feature $\mathbf{z}_m$ (\textit{resp.} $\mathbf{z}_c$). Afterward, $\mathbf{z}_m\dashrightarrow\mathbf{z}_c$ means that we use a variational distribution $p(\mathbf{z}_c|\mathbf{z}_m)$ to approximate the distribution $p(\mathbf{z}_c|\mathbf{I}_c)$. Here we denote this relationship using a dashed line, which is designed to be dynamic to diversify outputs.

\textbf{Generation.} In image completion, the images $\mathbf{I}_m$ and $\mathbf{I}_c$ can compose the image $\mathbf{I}_o$. Accordingly, for their latent features, we splice $\mathbf{z}_m$ and $\mathbf{z}_c$ to output $\mathbf{I}_o$. During training, we learn the decoder $g$ to finish the generation. 

\subsection{Image Completion Objective Decomposition}\label{sec:3.2}
We cannot observe the underlying $\mathbf{I}_c$ in the test procedure. Based on the probabilistic graph model, we perform the following variational inference with the Kullback-Leibler (KL) divergence: 
\begin{equation}\label{eq:variational}
\begin{aligned}
    &L_{\text{P}}=\text{KL}\left[q_{\psi}(\mathbf{I}_o,\mathbf{z}_m, \mathbf{z}_c|\mathbf{I}_m)\|p_{\phi}(\mathbf{I}_o,\mathbf{z}_m, \mathbf{z}_c|\mathbf{I}_m,\mathbf{I}_c)\right],\\
\end{aligned}
\end{equation}
where $q_{\psi}(\cdot|\cdot)$ and $p_{\phi}(\cdot|\cdot)$ are two conditioned distributions, with $\psi$ and $\phi$ being the parameters of their corresponding functions. The divergence~(\ref{eq:variational}) is minimized with respect to all parameters. It should be noted that, in fact, the overall objective should be: 
$$
\text{KL}[q_\psi(\mathbf{I}_o|\mathbf{I}_m)\|p_\phi(\mathbf{I}_o|\mathbf{I}_m,\mathbf{I}_c)]=\text{KL}[\mathbbm{E}_{\mathbf{z}_m,\mathbf{z}_c}[q_\psi(\mathbf{I}_o,\mathbf{z}_m,\mathbf{z}_c|\mathbf{I}_m)]\|\mathbbm{E}_{\mathbf{z}_m,\mathbf{z}_c}[p_\phi(\mathbf{I}_o,\mathbf{z}_m,\mathbf{z}_c|\mathbf{I}_m,\mathbf{I}_c)]].
$$
As the expectation in this formula is \textit{intractable}, we need to deal with the 1-step Monte Carlo sampling objective, which is why the divergence~(\ref{eq:variational}) is formed. Another advantage of using the divergence~(\ref{eq:variational}) is that constraining each sampling point can make the overall constraint tighter. We further decompose the divergence (\ref{eq:variational}) based on the probabilistic graph model as follows. 

\begin{proposition}
Regarding the KL divergence (\ref{eq:variational}), we show that the divergence can be decomposed as:
\begin{equation}\label{eq:analytical}
\begin{aligned}
    L_{\normalfont{{\text{P}}}}&=\normalfont{\text{KL}}\left[q_{\psi}(\mathbf{I}_o,\mathbf{z}_m, \mathbf{z}_c|\mathbf{I}_m)\|p_{\phi}(\mathbf{I}_o,\mathbf{z}_m, \mathbf{z}_c|\mathbf{I}_m,\mathbf{I}_c)\right]\\
    &=\underbrace{\mathbbm{E}_{(\mathbf{z}_m,\mathbf{z}_c)\sim q_{\psi}(\mathbf{z}_m, \mathbf{z}_c|\mathbf{I}_m)}\normalfont{\text{KL}}\left[q_{\psi}(\mathbf{I}_o|\mathbf{z}_m, \mathbf{z}_c)\|p_{\phi}(\mathbf{I}_o|\mathbf{I}_m,\mathbf{I}_c)\right]}_{\textcircled{a}}\\
    &+\underbrace{\mathbbm{E}_{\mathbf{z}_m\sim q_{\psi}(\mathbf{z}_m|\mathbf{I}_m)}\normalfont{\text{KL}}\left[q_{\theta}(\mathbf{z}_c|\mathbf{z}_m)\|p_{\phi}(\mathbf{z}_c|\mathbf{I}_c)\right]}_{\textcircled{b}}+\underbrace{\normalfont{\text{KL}}\left[q_{\psi}(\mathbf{z}_m|\mathbf{I}_m)\|p_{\phi}(\mathbf{z}_m|\mathbf{I}_m)\right]}_{\textcircled{c}},
\end{aligned}
\end{equation}
where $q_{\theta}$ is the variational distribution, and the parameters $\theta$ are involved in divergence minimization. 
\end{proposition}
The proof can be found in Appendix \ref{sec:A.1}. We analyze three terms in Eq.~(\ref{eq:analytical}) combining the probabilistic graph model and our purpose: 
\begin{itemize}[leftmargin=*]
    \item For \textcircled{a}, as discussed, we can generate $\mathbf{I}_o$ with $\mathbf{I}_m$ and $\mathbf{I}_c$, or with their latent features $\mathbf{z}_m$ and $\mathbf{z}_c$. We minimize the distance about the generation of $\mathbf{I}_o$ in the two ways. The minimization can ensure that the generation from the images and latent features is \textit{consistent}. 
    \item For \textcircled{b}, we do not limit the inference from $\mathbf{I}_c$ to $\mathbf{z}_c$ as a deterministic function. Instead, to diversify outputs, we use the variational distribution to restrain $q_{\theta}(\mathbf{z}_c|\mathbf{z}_m)$. The choice and discussion about the variational distribution will be provided in Section \ref{sec:3.3}. 
    \item For \textcircled{c}, it aims to keep that the inference using the variational distribution and the inference using the posterior is \textit{close}, which also guarantees the reliability of sampling. Besides, the variational distribution does not include $\mathbf{I}_c$ to enable latent features to be used in test. We employ the variational evidence lower bound (ELBO)  $L_{\text{ELBO}}$ \cite{kingma2013auto} to constrain \textcircled{c}. It is because, suppose that $p(\mathbf{I}_m)$ is a constant given $\mathbf{I}_m$, it is equivalent for  maximizing the $L_{\text{ELBO}}$ and minimizing \textcircled{c}. The details are provided in Appendix \ref{sec:A.4}. 
\end{itemize}

\subsection{Mixture-of-Experts Model}\label{sec:3.3}
As discussed, we need a suitable choice for the variational distribution $q_{\theta}(\mathbf{z}_c|\mathbf{z}_m)$ to output pluralistic results. In this work, we employ the Gaussian Mixture Model (GMM)~\cite{mclachlan2014number,mohri2018foundations} for the dynamic mode $q_{\theta}(\mathbf{z}_c|\mathbf{z}_m)$, which is more flexible than a unimodal and can diversify outputs. 

We denote the latent features of $\mathbf{I}_c$ that are inferred from $\mathbf{z}_m$ by $\hat{\mathbf{z}}_c$, which is distinguished from $\mathbf{z}_c$ achieved by the encoder. For the latent features $\mathbf{z}$, we denote the mean and covariance by $\bm{\mu}^{\mathbf{z}}$ and $\bm{\Sigma}^{\mathbf{z}}$. Mathematically, we have 
\begin{equation}\label{eq:gmm}
\begin{aligned}
q_{\theta}(\mathbf{z}_c|\mathbf{z}_m)=\sum_{i=1}^k \alpha_i\mathcal{N}\left(\mathbf{z}_c|\bm{\mu}_i^{\hat{\mathbf{z}}_c},\bm{\Sigma}_i^{\hat{\mathbf{z}}_c}\right),
\end{aligned}
\end{equation}
where $k$ denotes the number of primitives, $\alpha_i$ denotes the mixing coefficient of the $i$-th primitive, and $\bm{\mu}_i, \bm{\Sigma}_i$ denotes the distribution parameters of the $i$-th primitive. Note that the mixing coefficient $\alpha_i$ is the parameter of the Categorical distribution \cite{mohri2018foundations}. Intuitively, in Eq.~(\ref{eq:gmm}), we model $\mathbf{z}_m$ with GMM, where the $i$-th primitive can be used to represent $\hat{\mathbf{z}}_c^{(i)}$, \textit{i.e.},
$\bm{\mu}_i^{\hat{\mathbf{z}}_c}=\bm{\mu}_i^{\mathbf{z}_m}$ and $\bm{\Sigma}_i^{\hat{\mathbf{z}}_c}=\bm{\Sigma}_i^{\mathbf{z}_m}$. In this way, we can obtain different $\hat{\mathbf{z}}_c$ to meet diversity requirements. 

Besides, among all the parameters of GMM, except for $k$, the other parameters are called the \textit{inherent parameters} of GMM. 
In GMM (\ref{eq:gmm}), the number of primitives $k$ can be determined artificially according to the need for output diversity. The inherent parameters are optimized adaptively during training. Specifically, for the optimization of the mixing coefficients $\bm{\alpha}=[\alpha_1,\ldots,\alpha_k]$, we exploit the the \textit{frequency loss} \cite{ren2021probabilistic}. The objective is
\begin{equation}\label{eq:fre}
\begin{aligned}
&L_{\text{{F}}} = (\mathbf{v} - \bm{\alpha})(\mathbf{v} - \bm{\alpha})^\top, \mathbf{v}=[v_1,\ldots,v_k], \text{and}\\
&v_j = \Bigg\{
\begin{aligned}
&1, j=\argmin_i\normalfont{\text{KL}}\left[\mathcal{N}(\mathbf{z}_c|\bm{\mu}_i^{\hat{\mathbf{z}}_c},\bm{\Sigma}_i^{\hat{\mathbf{z}}_c})\|\mathcal{N}(\mathbf{z}_c|\bm{\mu}^{\mathbf{z}_c},\bm{\Sigma}^{\mathbf{z}_c})\right];\\
&0, \,\text{otherwise}.
\end{aligned}
\end{aligned}
\end{equation}
For our task, with the frequency loss (\ref{eq:fre}), the accumulated frequency approximate gradient is an \textit{asymptotically unbiased estimation} of the true gradient. We provide the detailed derivation and analysis in Appendix \ref{sec:A.2}. The reason we mention the frequency loss is that, the metric for evaluating the performance of each primitive is adapted to KL divergence in our task, which is different from the original metric in \cite{ren2021probabilistic}. For the optimization of $\bm{\mu}$ and $\bm{\Sigma}$ of $\mathbf{z}_m$, we utilize the \textit{back-propogate-max-operation} to reserve the distinguishable property of GMM (\ref{eq:gmm}). For our problem, the objective of the back-propogate-max-operation is shown in the following proposition.
\begin{proposition}\label{prop1}
By modeling the variational distribution with GMM (\ref{eq:gmm}), we have 
\begin{equation}\label{eq:kl_gmm}
\begin{aligned}
    L_{\normalfont{\text{BM}}}&
    =-\frac{1}{2}\log\frac{|\bm{\Sigma}_j^{\hat{\mathbf{z}}_c}|}{|\bm{\Sigma}^{\mathbf{z}_c}|}
    + \frac{1}{2}\normalfont{\text{tr}}\left({(\bm{\Sigma}^{\mathbf{z}_c})}^{-1}\bm{\Sigma}_j^{\hat{\mathbf{z}}_c}\right)-\frac{1}{2}(\bm{\mu}_j^{\hat{\mathbf{z}}_c} - \bm{\mu}^{\mathbf{z}_c})^\top{(\bm{\Sigma}^{\mathbf{z}_c})}^{-1}(\bm{\mu}_j^{\hat{\mathbf{z}}_c} - \bm{\mu}^{\mathbf{z}_c}),
\end{aligned}
\end{equation}
\end{proposition}
where $j=\argmin_i\normalfont{\text{KL}}\left[\mathcal{N}(\mathbf{z}_c|\bm{\mu}_i^{\hat{\mathbf{z}}_c},\bm{\Sigma}_i^{\hat{\mathbf{z}}_c})\|\mathcal{N}(\mathbf{z}_c|\bm{\mu}^{\mathbf{z}_c},\bm{\Sigma}^{\mathbf{z}_c})\right]$, and $\text{tr}(\cdot)$ denotes the trace of a matrix. The derivation of Proposition \ref{prop1} can be found in Appendix \ref{sec:A.3}. For Proposition \ref{prop1}, if we minimize $L_{\normalfont{\text{BM}}}$, we optimize $\bm{\mu}$ and $\bm{\Sigma}$ of $\mathbf{z}_m$, which reduces the distance between $\hat{\mathbf{z}}_c^{(j)}$ and $\mathbf{z}_c$. For the term \textcircled{b} in Eq.~(\ref{eq:analytical}), we then have $L_{\text{GMM}} = \mathbbm{E}_{\mathbf{z}_m\sim q_{\psi}(\mathbf{z}_m|\mathbf{I}_m)}\normalfont{\text{KL}}\left[q_{\theta}(\mathbf{z}_c|\mathbf{z}_m)\|p_{\phi}(\mathbf{z}_c|\mathbf{I}_c)\right] \approx L_{\text{F}} + L_{\text{BM}}$.  

\textbf{Discussion.} Note that the diversity of results is directly related to $\mathbf{z}_m$. When GMM for $\mathbf{z}_m$ is \textit{dominated} by the $i$-th primitive ($\alpha_i\approx1$), the diversity of image completion results is reduced. Although this phenomenon did not appear in our experiments, it may exist in practice. Actually, this phenomenon is reasonable. From human cognition, if $\mathbf{I}_m$ shows \textit{only one pattern}, the inference for $\mathbf{I}_c$ will be deterministic \cite{bertalmio2000image}. For example, if the size of the missing region of an incomplete image is very small, we can infer the missing region both deterministically and reasonably, where the diversity of results is reduced. Prior methods with task-agnostic constraints do not consider this fact, which may create \textit{unreasonable results}. For our method, the constraints for diversity are task-related, which can take this phenomenon into account. 
\begin{figure*}[!t]
    \centering
    \includegraphics[width=0.80\textwidth]{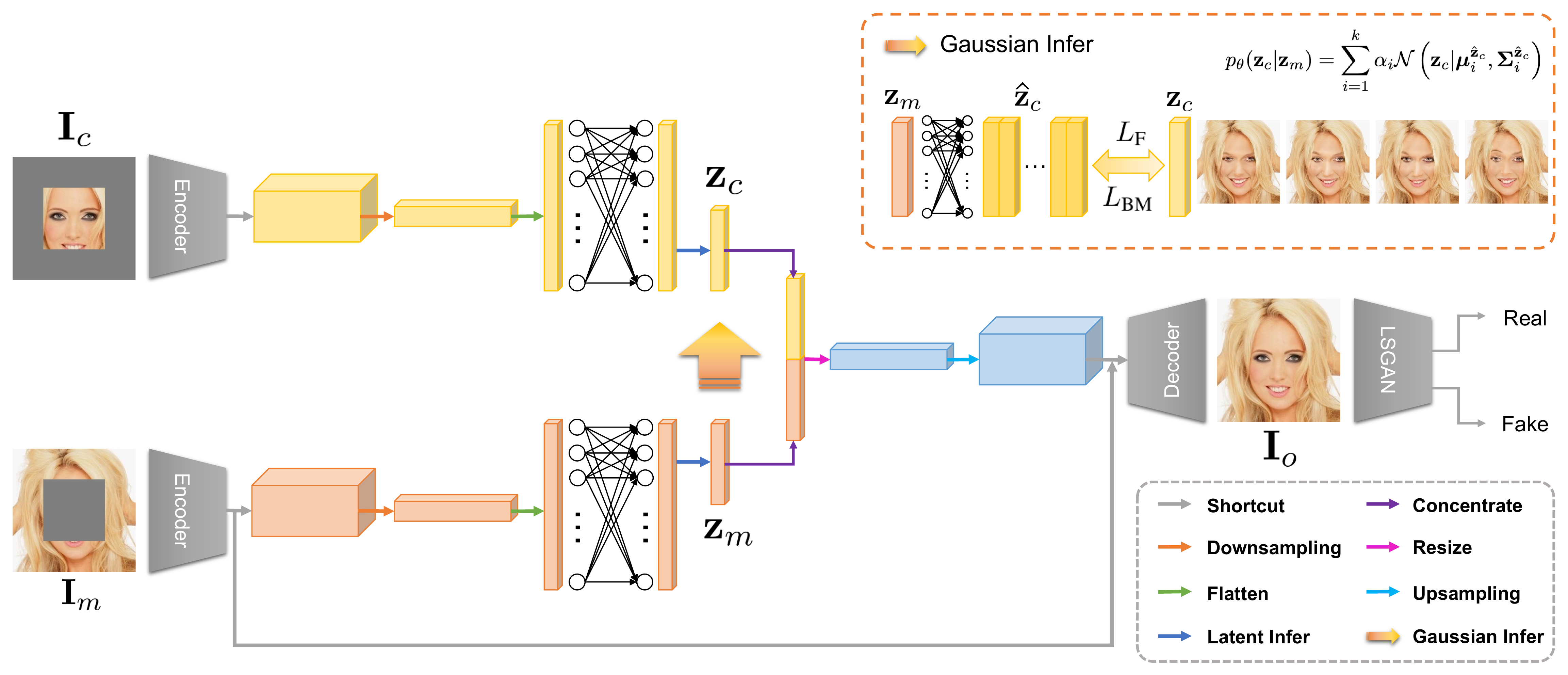}
    \caption{The algorithm framework of the proposed method.}
    \label{fig:fw}
\end{figure*}

\subsection{Reconstruction and Adversarial Losses}\label{sec:3.4}
To ensure the minimization of the term \textcircled{a} in Eq.~(\ref{eq:analytical}), we incorporate the use of reconstruction and adversarial losses \cite{goodfellow2014generative}. The reconstruction loss is formulated as: 
\begin{equation}\label{eq:recon}
    L_{\text{R}}=\mathbbm{E}\left[\|\hat{\mathbf{I}}_o-\mathbf{I}_{o}\|_1\right]+\mathbbm{E}\left[\|\hat{\mathbf{I}}^{(j)}_m-\mathbf{I}_{m}\|_1\right],
\end{equation}
where $\hat{\mathbf{I}}_o$ is generated from $\mathbf{z}_m$ and $\mathbf{z}_c$, and $\hat{\mathbf{I}}_m^{(j)}$ is generated from $\hat{\mathbf{z}}^{(j)}_c$. Here, $\hat{\mathbf{z}}^{(j)}_c$ can be inferred from $\mathbf{z}_m$ by using Eq.~(\ref{eq:gmm}) and Eq.~(\ref{eq:fre}). The image $\hat{\mathbf{I}}_m^{(j)}$ is obtained from $\hat{\mathbf{I}}_o^{(j)}$ with the image mask that can degenerate $\mathbf{I}_o$ to $\mathbf{I}_m$. The image $\hat{\mathbf{I}}_o^{(j)}$ is generated from $\mathbf{z}_m$ and $\hat{\mathbf{z}}^{(j)}_c$. The reconstruction loss (\ref{eq:recon}) controls the visual rationality of image completion results. Also, we efficiently avoid the deterministic fashion with GMM. Furthermore, we leverage adversarial training \cite{goodfellow2014generative} to make the generated images more realistic. The adversarial loss $L_{\text{A}}$ is formulated as 
\begin{equation}
    L_{\text{A}}=\mathbbm{E}\left[\|\mathcal{D}(\hat{\mathbf{I}}_o) - 1\|_2^2\right]+\mathbbm{E}\left[\|\mathcal{D}(\hat{\mathbf{I}}^{(j)}_o) - \mathcal{D}(\mathbf{I}_{o})\|_2^2\right], 
\end{equation}
where $\mathcal{D}$ is the discriminator optimized by the discriminator loss based on LSGAN \cite{mao2017least}. We jointly train our encoder $f$ and decoder $g$ through the following combined loss: 
\begin{equation}\label{eq:combined_loss}
    L_\text{C}=L_\text{R}+\lambda_\text{A} L_\text{A},
\end{equation}
where the weight $\lambda_\text{A}$ is set to 0.05 in all experiments. Then, the final objective of the loss function is the sum of three losses, \textit{i.e.}, $L_{\rm{FINAL}}=L_{\rm{GMM}}+L_{\rm{ELBO}}+L_{\rm{C}}$. The final loss is to maximize the log-likelihood of the conditional data distribution. Specifically, the adversarial loss uses the Wasserstein distance to make generated images realistic given latent features. The ELBO loss uses KL divergence and distribution log-likelihood to make variational posterior approximate real posterior. 

\textbf{Algorithm Flows.} For the convenience of following technical details, we provide the algorithm flows. The illustration is shown in Figure \ref{fig:fw}. The algorithm flows of training and test stages can be found in Algorithm \ref{alg:train} and Algorithm \ref{alg:test}. 

It should be noted that the number of primitives $k$ does not mean that we are limited to generate only $k$ diverse results given $\mathbf{I}_m$ (in Steps 3 and 4 of Algorithm \ref{alg:test}). In fact, we can sample from the $k$ primitives to obtain lots of image completion results. Moreover, as GMM has a much larger capacity than the unimodal Gaussian distribution \citep{mohri2018foundations}, we can achieve greater diversity. We provide empirical observations in Section \ref{sec:4.3} and Appendix \ref{sec:C}.

\begin{algorithm}[!t]
\caption{The training procedure of our method}\label{alg:train}

\textbf{Input:} images $\mathbf{I}_o$, $\mathbf{I}_m$, and $\mathbf{I}_c$, the fixed number of primitives of GMM $k$, the initialized encoder $f$, and decoder $g$. 

\begin{algorithmic}[1]
\STATE \textbf{Encode} $\mathbf{I}_m$ (\textit{resp}. $\mathbf{I}_c$) to $\mathbf{z}_m$ (\textit{resp}. $\mathbf{z}_c$) with $f$;
\STATE \textbf{Model} $\mathbf{z}_m$ with GMM and $\mathbf{z}_c$ with a unimodal Gaussian distribution;
\STATE \textbf{Calculate} the loss $L_{\normalfont{\text{GMM}}}$ as discussed in Section \ref{sec:3.3};
\STATE \textbf{Update} the parameters of GMM with $L_{\normalfont{\text{GMM}}}$; 
\STATE \textbf{Infer} $\hat{\mathbf{z}}_c^{(j)}$ from $\mathbf{z}_m$ with  Eq.~(\ref{eq:gmm}) and Eq.~(\ref{eq:fre}), $j=1,\ldots,k$; 
\STATE \textbf{Generate} images by $g$ with $\hat{\mathbf{I}}_o = g(\mathbf{z}_m, \mathbf{z}_c)$ and $\hat{\mathbf{I}}_o^{(j)} = g(\mathbf{z}_m, \hat{\mathbf{z}}^{(j)}_c)$, $j=1,\ldots,k$;
\STATE \textbf{Calculate} the loss $L_{\text{ELBO}}$ and $L_\text{C}$ as discussed in Section \ref{sec:3.2} and \ref{sec:3.4};
\STATE \textbf{Update} all the parameters \textit{w.r.t.} $L_{\text{ELBO}}$ and $L_\text{C}$.
\end{algorithmic}
\textbf{Output:} the trained encoder $f^*$ and decoder $g^*$. 
\end{algorithm}

\vspace{-10pt}
\begin{algorithm}[!t]
\caption{The test procedure of our method}\label{alg:test}
\textbf{Input:} the image $\mathbf{I}_m$, the fixed number of primitives of GMM $k$, the trained encoder $f^*$, and decoder $g^*$. 

\begin{algorithmic}[1]
\STATE \textbf{Encode} $\mathbf{I}_m$ to $\mathbf{z}_m$ with $f^*$;
\STATE \textbf{Model} $\mathbf{z}_m$ with GMM;
\FOR{$j=1$ to $k$}
\STATE \textbf{Sample} $i$ with $i \sim \texttt{Categorical}(\alpha_1, \ldots, \alpha_k)$;
\STATE \textbf{Infer} $\hat{\mathbf{z}}^{(j)}_c$ with $\hat{\mathbf{z}}^{(j)}_c=\mathcal{N}(\bm{\mu}_i^{\mathbf{z}_m},\bm{\Sigma}_i^{\mathbf{z}_m})$;
\STATE \textbf{Generate} images $\hat{\mathbf{I}}_o^{(j)}$ by $g^*$ with $\hat{\mathbf{I}}_o^{(j)} = g^*(\mathbf{z}_m, \hat{\mathbf{z}}^{(j)}_c)$.
\ENDFOR
\end{algorithmic}
\textbf{Output:} pluralistic image completion results $\{\hat{\mathbf{I}}_o^{(j)}\}_{j=1}^k$.
\end{algorithm}

\section{Experiments}\label{sec:4}

In this section, we conduct a series of experiments to justify our claims. We first introduce the implementation of our method (Section \ref{sec:4.1}). The comprehensive experimental results and comparison with advanced methods are then provided and discussed (Section \ref{sec:4.2}). Finally, we conduct an analysis study to present and discuss our method in more detail (Section \ref{sec:4.3}).

\subsection{Implementation Details}\label{sec:4.1}
\textbf{Datasets.} We evaluated our proposed model on five popularly used datasets, \textit{i.e.}, CelebA-HQ \cite{karras2017progressive,liu2015deep}, FFHQ \cite{karras2019style}, Paris StreetView \cite{doersch2012makes}, Places2 \cite{zhou2017places}, and ImageNet \cite{russakovsky2015imagenet}. We verify the effectivness of our method with different types of mask regions, including both center and random masks \cite{zheng2019pluralistic,wan2021high}. 

\textbf{Network and Optimization.} The proposed method can be implemented efficiently. The encoder and decoder of our pipeline are inspired by PIC \cite{zheng2019pluralistic} for comparison. We apply average pooling and interpolation with convolutional layers to implement downsampling and upsampling respectively. During optimization, we use the Adam optimizer \cite{kingma2015adam}. The learning rate is fixed to $10^{-4}$ during the training procedure.   

\textbf{Baselines.} We compare the proposed method with the following state-of-the-art methods, which include (1) single image completion methods: DFv2 \cite{yu2019free}, EC \cite{nazeri2019edgeconnect}, and MED \cite{liu2020rethinking}; (2) pluralistic image completion methods: PIC \cite{zheng2019pluralistic} and ICT \cite{wan2021high}. We abbreviate our method (\textbf{P}luralistic \textbf{I}mage \textbf{C}ompletion with Probabilistic \textbf{M}ixture-of-\textbf{E}xperts) as PICME. The methods are implemented by PyTorch and evaluated on NVIDIA Tesla A100 GPUs. 

\textbf{Measurement.} For measurement, we provide both qualitative and quantitative results. For the metrics of quantitative results, we use common evaluation metrics such as the peak signal-to-noise ratio (PSNR), structural similarity (SSIM), mean absolute error (MAE), and fr$\acute{\text{e}}$chet inception distance (FID) \cite{heusel2017gans} to measure the similarity between the image completion result and ground truth. Furthermore, pluralistic image completion is supposed to focus on generating diverse realistic results rather than merely approximating ground-truth ones \cite{zhao2020uctgan}. To \textit{measure the \textit{result diversity}}, we add two perceptual quality metrics to quantitative results, which include LPIPS \cite{zheng2019pluralistic} and DIV-FID \cite{wan2021high}. 

\begin{figure}[!t]
    \centering
    \includegraphics[width=0.65\textwidth]{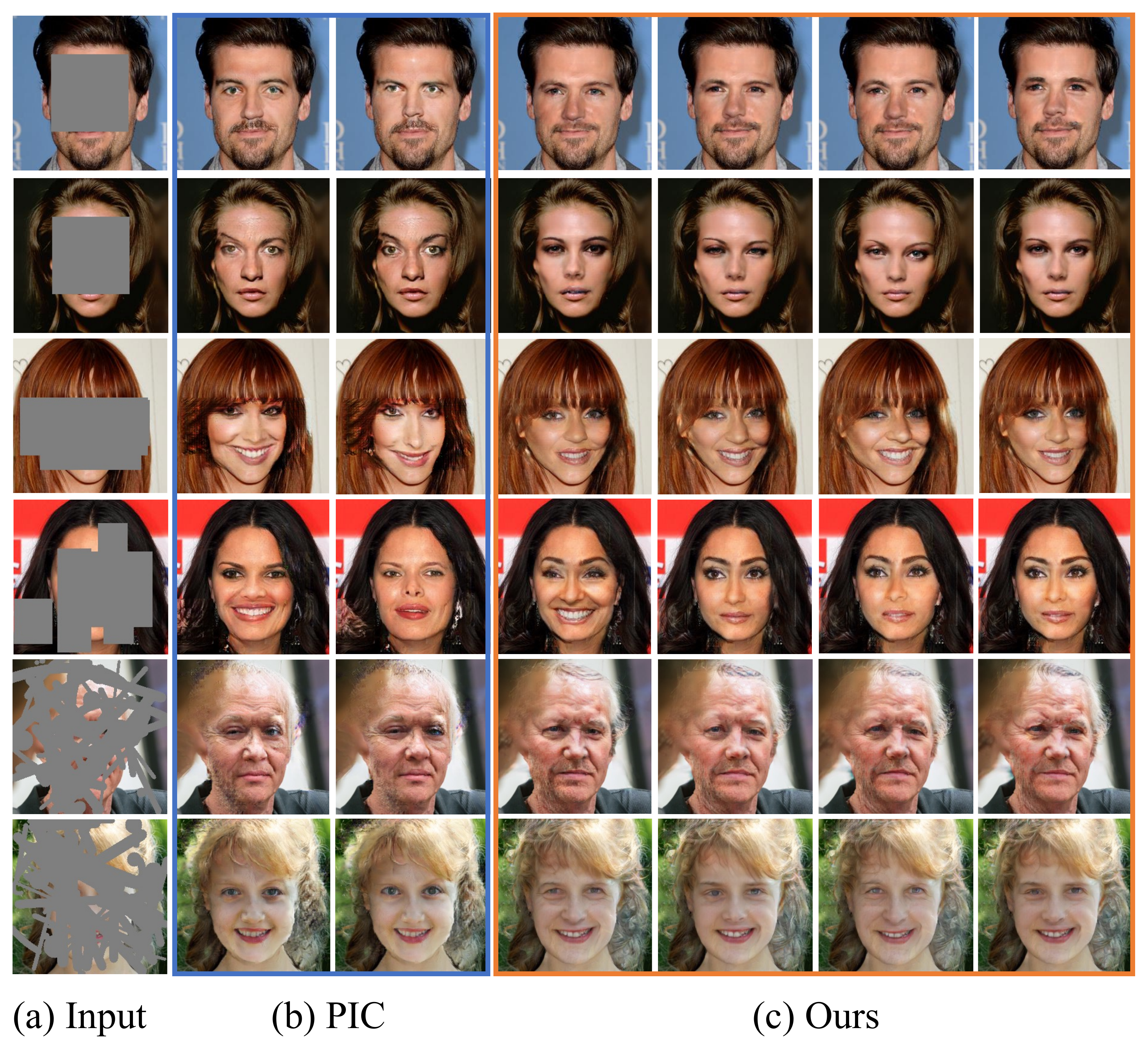}
    \caption{Qualitative comparison of our method with PIC  on CelebA-HQ (\textbf{first four rows}) and FFHQ (\textbf{last two rows}). Best viewed by zooming in.}
    \label{fig:celeba1}
\end{figure}

\begin{figure}[!t]
    \centering
    \includegraphics[width=0.65\textwidth]{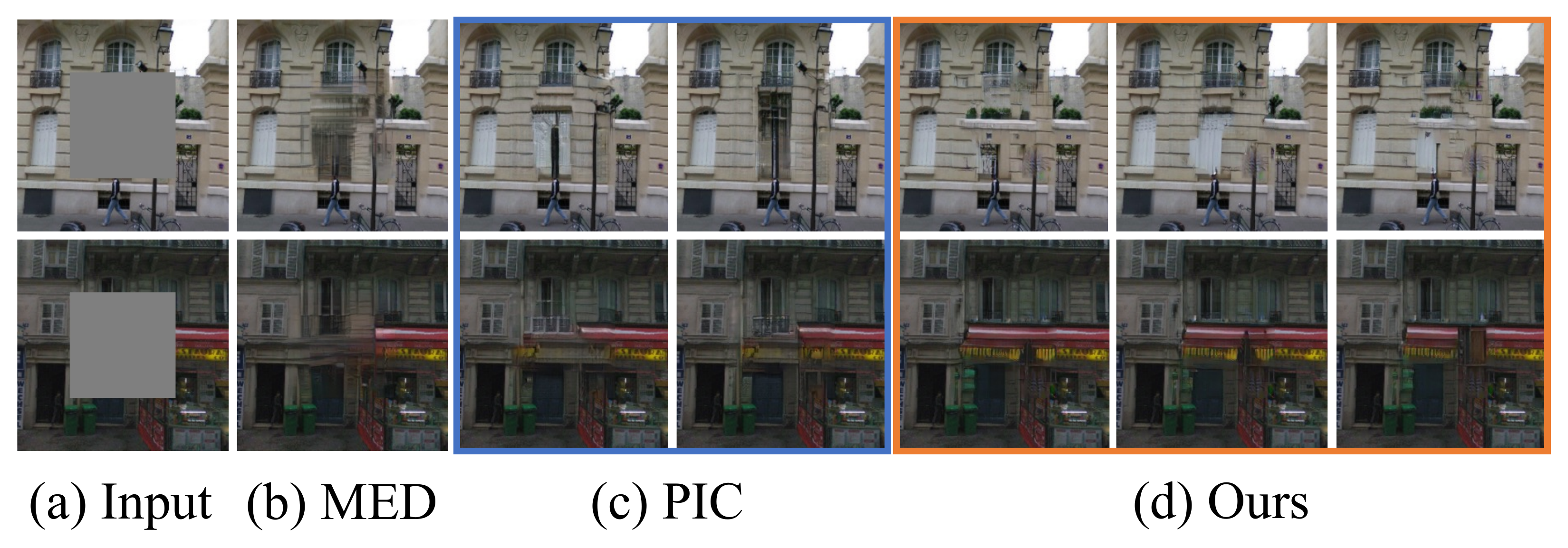}
    \caption{Qualitative comparison of our method with MED and PIC on Paris StreetView. Best viewed by zooming in.}
    \label{fig:paris1}
\end{figure}

\begin{figure}[!t]
    \centering
    \includegraphics[width=0.65\textwidth]{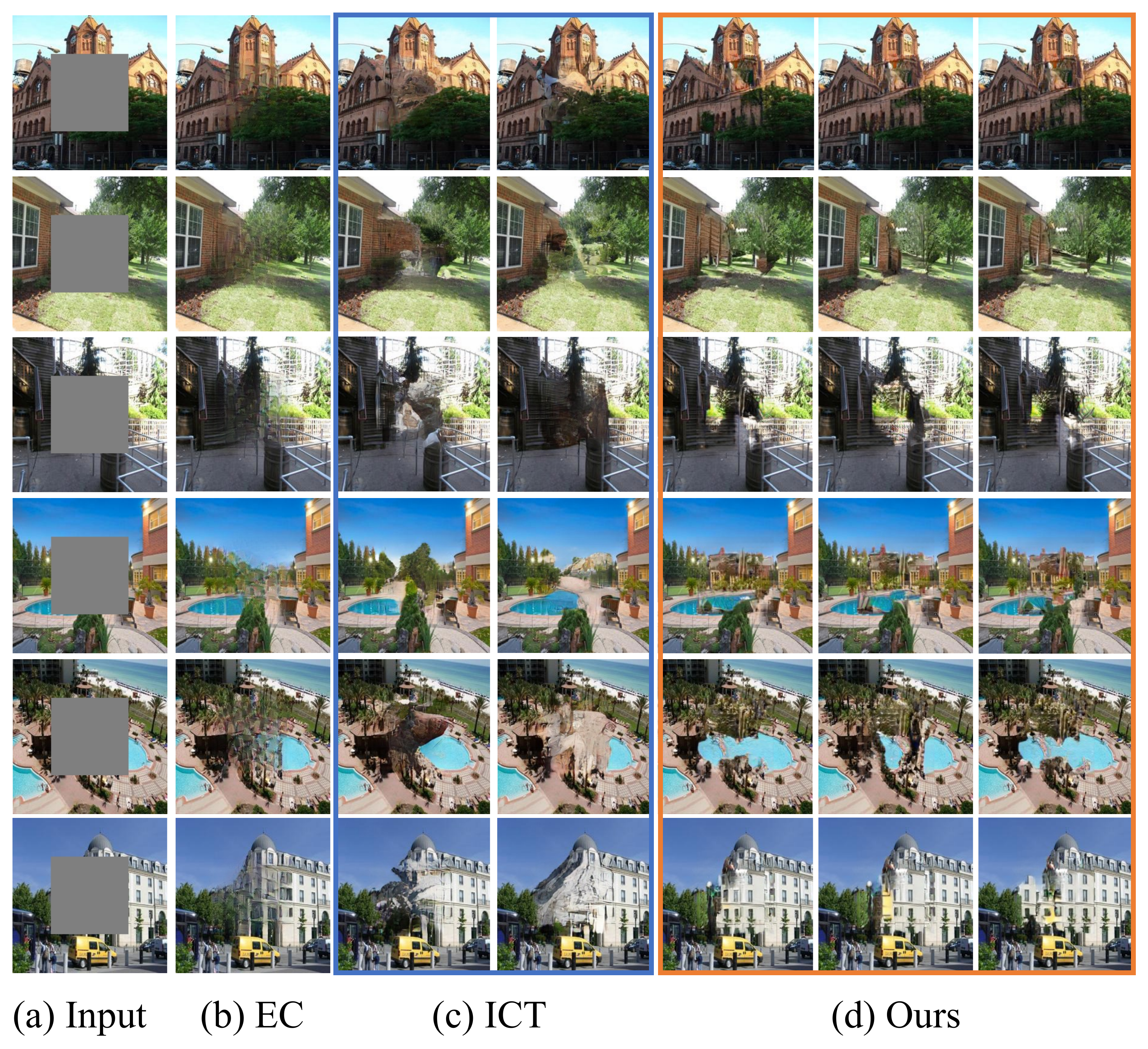}
    \caption{Qualitative comparison of our method with EC and ICT on Places2. Best viewed by zooming in.}
    \label{fig:places}
\end{figure}

\begin{figure}[!t]
    \centering
    \includegraphics[width=0.65\textwidth]{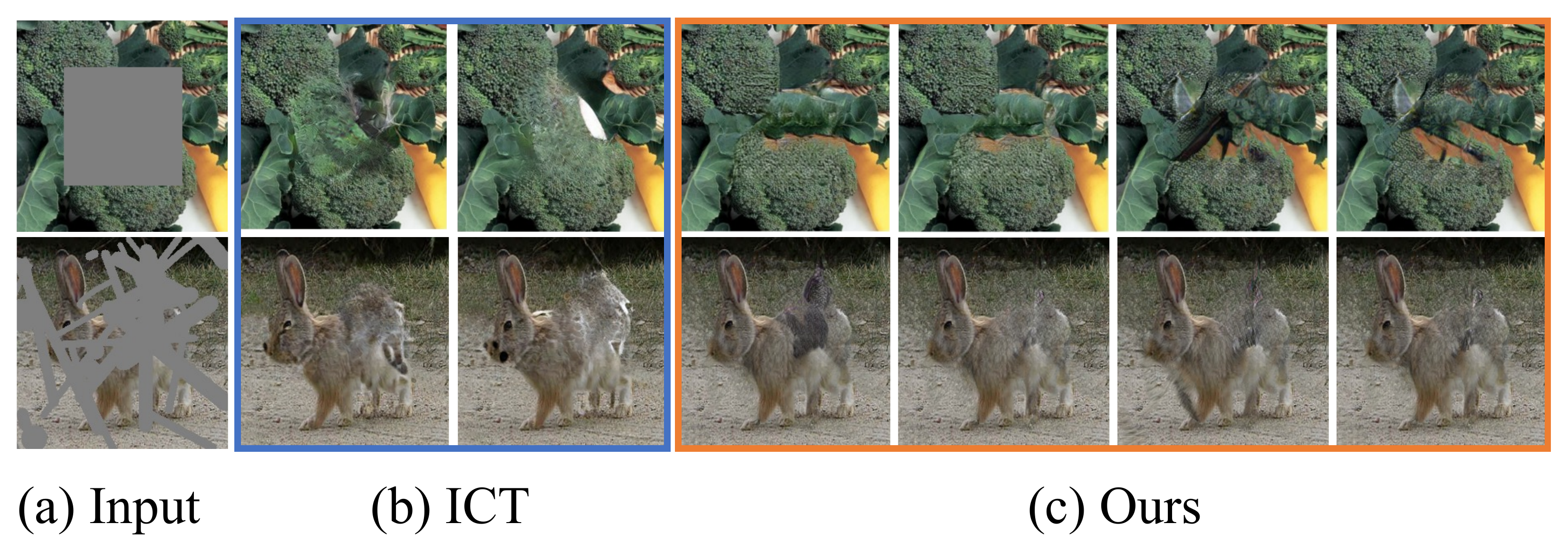}
    \caption{Qualitative comparison of our method with ICT on ImageNet. Best viewed by zooming in.}
    \label{fig:imagenet}
\end{figure}

\begin{figure}[!h]
    \centering
    \includegraphics[width=0.55\textwidth]{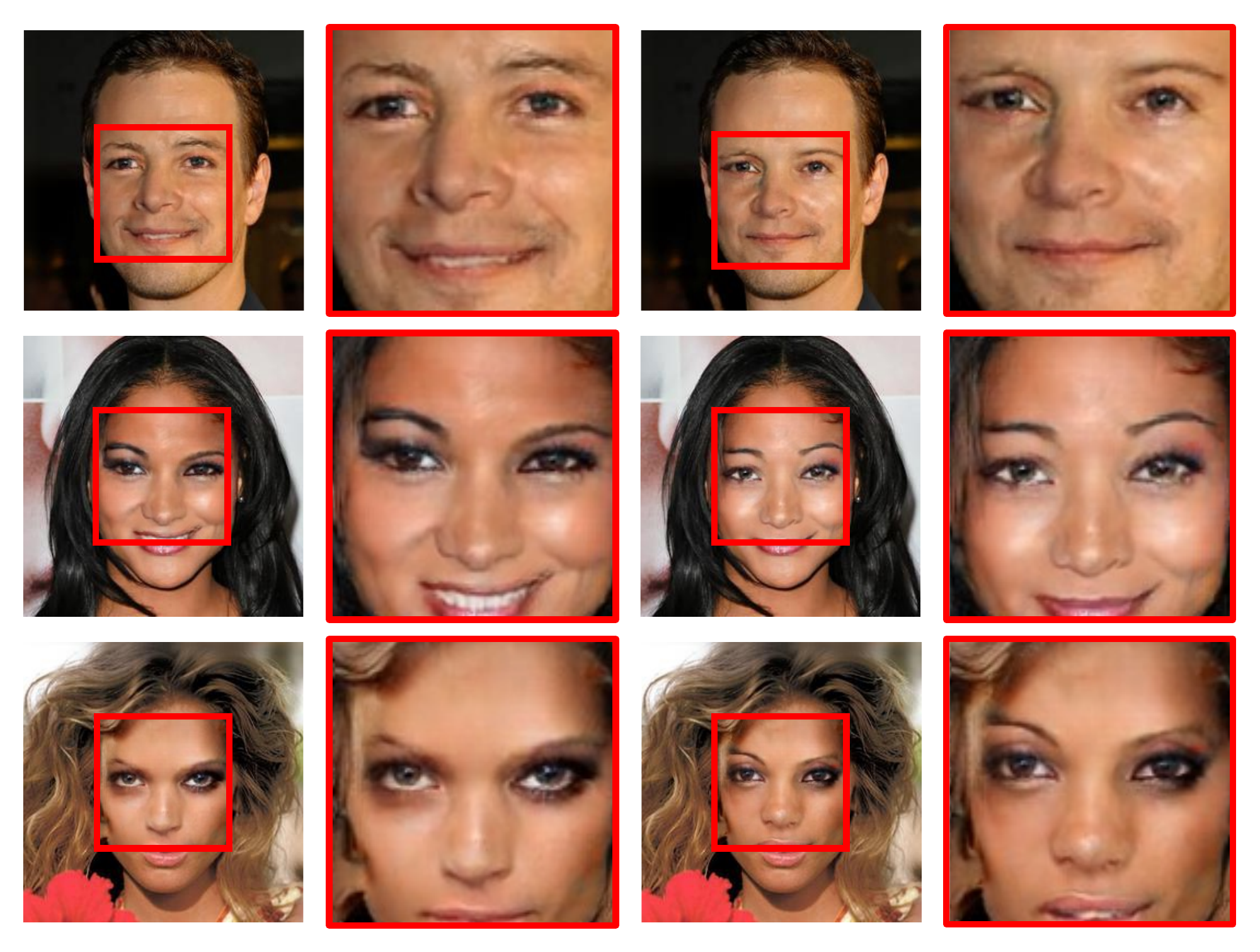}
    \caption{Refined result analysis of our method. The images come from CelebA-HQ.}
    \label{fig:details}
\end{figure}

\subsection{Comparison with Prior Methods}\label{sec:4.2}
\subsubsection{Qualitative Comparisons.}
We provide extensive qualitative comparison results to justify our claims. First, we show the results on  CelebA-HQ and FFHQ in Figure~\ref{fig:celeba1}, which are images related to human faces. As can be seen, the image completion results achieved by our method are realistic, following high image quality. Also, compared with the baseline PIC, our completion results are more diverse, \textit{e.g.}, see the images in the second and last rows.

Second, we provide the results on Paris StreetView and Places in Figures~\ref{fig:paris1} and \ref{fig:places} respectively. For the results on Paris StreetView, we can see that image completion by the baselines MED and PIC are somewhat distorted. By contrast, our method can finish the image completion task better. The details can be checked in the complements to walls. Then, we turn the attention to the image completion results on Places2. Our method still achieves superior performance compared with baselines.

Lastly, we provide the results on ImageNet in Figure \ref{fig:imagenet}. Note that the sources of the data in ImageNet are very complicated. The image completion task is rather challenging on this dataset, even for single image completion \cite{zheng2019pluralistic}. Different from some prior work \cite{pathak2016context,iizuka2017globally} that were trained on a 100k subset of training images of ImageNet, we directly the network on the original ImageNet training dataset with all images. The results mean that our method can infer the content effectively.

\textbf{Refined Result Analysis.} We provide some refined results in Figure~\ref{fig:details} to analyze the image completion details of our method. For the face images, we can find that the image completion results have very different facial expressions.

\begin{table*}[!t]
\small
\begin{center}
    \begin{tabular}{l|c|cccc|cccc}
    \toprule
        \multicolumn{2}{c}{Dataset}           & \multicolumn{4}{c}{FFHQ} & \multicolumn{4}{c}{Places2}      \\
\hline
Method & Mask           & PSNR $\uparrow$   & SSIM $\uparrow$  & MAE $\downarrow$    & FID $\downarrow$ & PSNR $\uparrow$   & SSIM $\uparrow$  & MAE $\downarrow$    & FID $\downarrow$    \\
\hline
DFv2   & \multirow{6}{*}{Center} & 25.868 & 0.922 & 0.0231 & 16.278 & 26.533 & 0.881 & 0.0215 & 24.763  \\
EC     &            &26.901 & 0.938 & 0.0209 & 14.276                  & 26.520 & 0.880 & 0.0220 & 25.642  \\
MED    &            & 26.325 & 0.922 & 0.0230 & 14.791                  & 26.469 & 0.877 & 0.0224 & 26.977 \\
PIC    &             & 26.781& 0.933 & 0.0215 & 14.513                 & 26.099 & 0.865 & 0.0236 & 26.393  \\
ICT    &            & \textbf{27.922} & \textbf{0.948} & 0.0208 & 10.995                  & 26.503 & 0.880 & 0.0244 & \textbf{21.598}  \\
PICME\textsuperscript{\dag}   &            & 27.551 & 0.937 & \textbf{0.0203} & \textbf{10.604}                   & \textbf{26.554}      & \textbf{0.886}     & \textbf{0.0208}      & 24.373     \\
\hline
DFv2   & \multirow{6}{*}{Random}   & 24.962 & 0.882 & 0.0310 & 19.506   & 25.692 & \textbf{0.834} & 0.0280 & 29.981  \\
EC     &                & 25.908 & 0.882 & 0.0301 & 17.039              & 25.510 & 0.831 & 0.0293 & 30.130 \\
MED    &     & 25.118 & 0.867 & 0.0349 & 19.644                         & 25.632 & 0.827 & 0.0291 & 31.395 \\
PIC    &           & 25.580 & 0.889 & 0.0303 & 17.364                   & 25.035 & 0.806 & 0.0315 & 33.472 \\
ICT    &              & \textbf{26.681} & 0.910 & 0.0292 & \textbf{14.529}                & \textbf{25.788} & 0.832 & 0.0267 & 25.420 \\
PICME\textsuperscript{\dag}   &         & 25.950 & \textbf{0.912} & \textbf{0.0289} & 17.014                     & 25.310      & 0.829     & \textbf{0.0236}      & \textbf{25.025} \\
\bottomrule
    \end{tabular}
    \caption{Quantitative results on FFHQ and Places2 datasets with different mask settings. The best results are in \textbf{bold}.}
    \label{tab:place2_ffhq}
    \vspace{-10pt}
    \end{center}
\end{table*}

\begin{table}[!t]
\begin{center}
    \begin{tabular}{l|c|cccc}
    \toprule
        \multicolumn{2}{c}{Dataset}           & \multicolumn{4}{c}{ImageNet}       \\
\hline
Method & Mask            & PSNR $\uparrow$   & SSIM $\uparrow$  & MAE $\downarrow$    & FID $\downarrow$   \\
\hline
PIC   & \multirow{3}{*}{Center} &  24.010 & 0.867 & 0.0319 & 47.750\\
ICT    &                              &  24.757 & 0.888 & 0.0263 & 28.818 \\
PICME\textsuperscript{\dag}   &                              & \textbf{24.932}      & \textbf{0.897}     & \textbf{0.0260}      & \textbf{23.718}      \\
\hline
PIC   & \multirow{3}{*}{Random}      &  22.711 & 0.791 & 0.0462 & 59.428\\
ICT    &                              & \textbf{23.775} & 0.835 & \textbf{0.0358} & \textbf{35.842}\\
PICME\textsuperscript{\dag}   &                              & 23.322      & \textbf{0.846}     & 0.0415      & 39.742 \\
\bottomrule
    \end{tabular}
    \caption{Quantitative results on ImageNet with different mask settings. The best results are in \textbf{bold}.}
    \label{tab:imagenet}
     \vspace{-10pt}
    \end{center}
\end{table}

\begin{table*}[!t]
\begin{center}
    \begin{tabular}{l|cc|cc}
    \toprule
        \multicolumn{1}{c}{Dataset}           & \multicolumn{2}{c}{FFHQ} & \multicolumn{2}{c}{Places2}      \\
\hline
Method           & LPIPS $\uparrow$   & DIV-FID $\uparrow$  & LPIPS $\uparrow$   & DIV-FID $\uparrow$      \\
\hline
PIC   & 0.029 & 9.130 & 0.047 & 17.742   \\
ICT   & 0.065 & 13.909 & 0.089 & 25.253   \\
PICME\textsuperscript{\dag} & \textbf{0.071}      & \textbf{17.361}  &  \textbf{0.092}      & \textbf{27.194}           \\
\bottomrule
    \end{tabular}
    \caption{The diversity result comparison on FFHQ and Places2 datasets with the center mask setting. The best results are in \textbf{bold}.}
    \label{tab:div}
     \vspace{-10pt}
    \end{center}
\end{table*}

\begin{table*}[!t]
\begin{center}
    \begin{tabular}{l|c}
    \toprule
Method           & Inference time (minutes) $\downarrow$        \\
\hline
PIC    & 0.387 \\
ICT   & 179.126  \\
PICME\textsuperscript{\dag}  & \textbf{0.235}        \\
\bottomrule
    \end{tabular}
    \caption{The inference time comparison with PIC and ICT in the center mask setting. The best result is in \textbf{bold}.}
    \label{tab:infertime}
     \vspace{-10pt}
    \end{center}
\end{table*}

\subsubsection{Quantitative Comparisons}
\textbf{Common Metrics.} Quantitative evaluation with common metrics is difficult for pluralistic image completion, as our goal is to obtain diverse but reasonable solutions for a masked image \cite{zheng2019pluralistic}. To make a feasible comparison, as did in \cite{zheng2019pluralistic}, we sample one completed image many times from $k$ images to compare with the original image. The results are provided in Tables~\ref{tab:place2_ffhq} and \ref{tab:imagenet}. As can be seen, for FFHQ and Places2, on both the settings of center and random masking, our method achieves superior performance. On ImageNet, our method performs the best with the center masking consistently, while it is competitive with the random masking. 

It should be noted that ICT is a very strong baseline and works better than our method in some cases. It is because ICT is built with the image transformer \cite{parmar2018image}, which is powerful in vision tasks. Compared with ICT, the advantages of our method lies in not only better interpretability as discussed, but also faster inference speeds. We will demonstrate this in inference time.

\textbf{Diversity Metrics.} For pluralistic image completion, the diversity evaluations are significant. We generate five completion images in each task and report the mean of the diversity metric. The results are shown in Table \ref{tab:div}, which represents that our method can generate more diverse completion images than baselines. The results with common and diversity metrics support our claim very well. That is to say, the image completion results of our method are both high-quality and pluralistic.

\textbf{Inference Time.} We report the inference time of our method, compared with PIC and ICT. All inference runs on one NVIDIA Tesla A100 GPU for fairness. The size of images is $256\times 256$. We perform pluralistic image completion for 100 different images. For each image, we generate 6 completion results for it. We report the total inference time for these images. The results are shown in Table \ref{tab:infertime}. Note that as ICT is a transformer-based method, the calculated consumption of its inference is much heavier. What's worse, ICT needs \textit{iterative Gibbs sampling} during inference. The two issues make ICT have to face heavy computational consumption, which is also mentioned in \cite{wan2021high}. In contrast, our method is more inference-efficiency, which is \textbf{more than 750 times faster than ICT}. The merit could make our method easier to use in the real world.

\subsection{More Analyses and Justifications}\label{sec:4.3}
\begin{figure}[!t]
    \centering
    \includegraphics[width=0.65\textwidth]{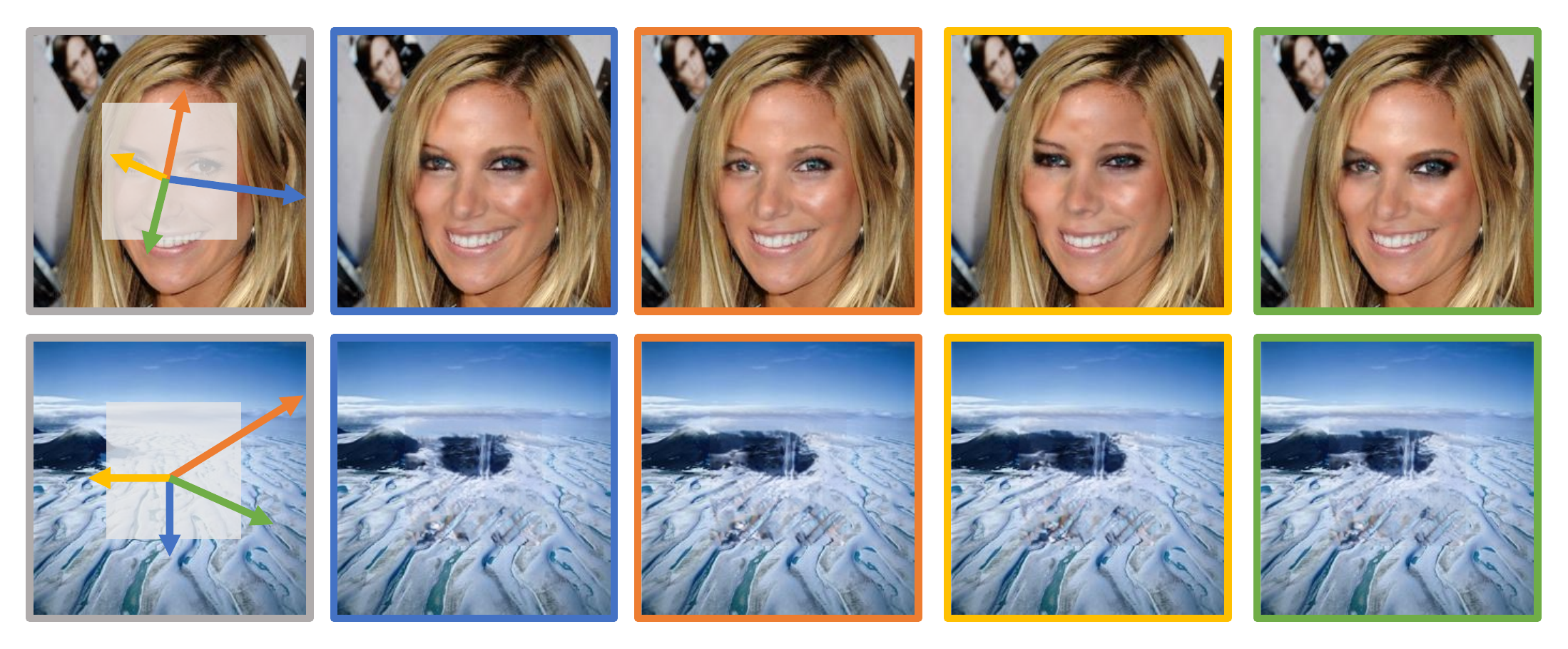}
    \caption{GMM primitive visualization of pluralistic completion results by our method. The face image comes from CelebA-HQ. The scene image comes from Places2.}
    \label{fig:arrow1}
\end{figure}

\begin{figure}[!h]
    \centering
    \includegraphics[width=0.65\textwidth]{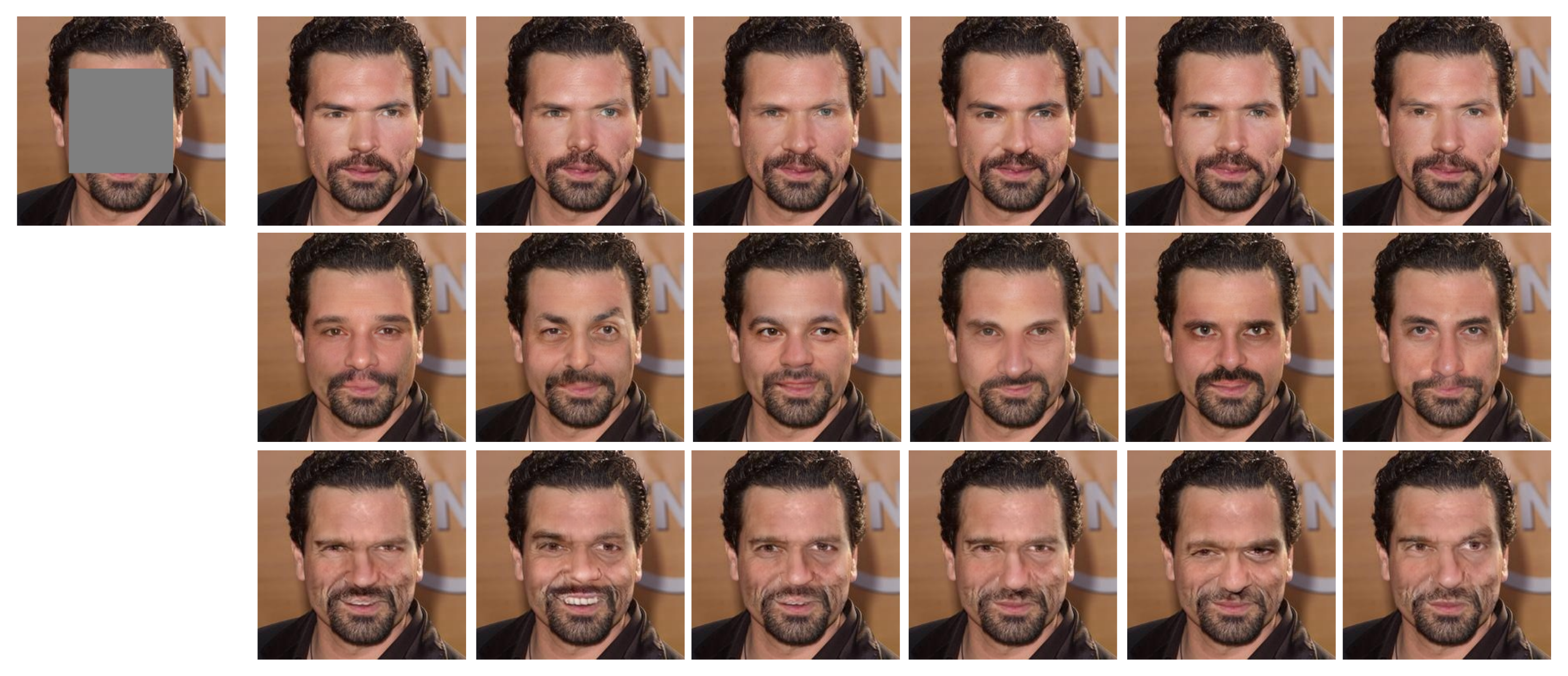}
    \caption{Refined diversity comparison with PIC (\textbf{the first row}) and ICT (\textbf{the second row}). The original image comes from CelebA-HQ.}
    \label{fig:gmmcom}
\end{figure}

\textbf{GMM Primitive Visualization.} We further stress the diversity of the completion results achieved by our method. Specifically, we visualize the $\hat{\mathbf{z}}_c$ which is inferred from $\mathbf{z}_m$ as discussed. The visualization results are obtained with t-SNE \cite{van2008visualizing}. Different $\hat{\mathbf{z}}_c$ are presented by the 2D-vectors in different colors, which are shown in Figure~\ref{fig:arrow1}. As can be seen, the vectors are \textit{scattered}, which clearly demonstrates that the completion results of our method are diverse. 

\textbf{Refined Diversity Comparison.} We argue that, benefiting from that GMM has larger capacities than a unimodal Gaussian distribution, the completion results of our method would be more diverse. To justify our claims, we provide the refined diversity comparison of our method with PIC and ICT, as shown in Figure~\ref{fig:gmmcom}. Note that the six image completion results of PIC are obtained by sampling from one unimodal Gaussian distribution. The results of our method are obtained by sampling from six primitives. Clearly, our results are more diverse than PIC's. Besides, for ICT, it does not explicitly control the output diversity. The diversity of pluralistic image completion is thus \textit{less interpretable} than our method. 

In addition, we state that our method is not limited to only generating $k$ completion results before. In fact, we can sample from $k$ primitives of GMM to obtain lots of completion results. Due to the limited page, we present the results and more analyses in Appendix \ref{sec:C}. Moreover, the richer completion result comparison is presented in Appendix \ref{sec:D}. 

\section{Conclusion}\label{sec:5}
In this paper, we focus on the complicated and challenging problem of pluralistic image completion. We propose a novel end-to-end probabilistic method for this challenging problem. Based on the probabilistic graph model, our method divides the entire procedure of pluralistic image completion into several sub-procedures, where GMM is used to diversify outputs. Experiments on a variety of datasets show that the image completion results by our method are both high-quality and pluralistic. In future work, investigating the feasibility of the proposed method for other diverse decision-making scenarios might prove important.

\section*{Acknowledgments}\label{sec:6}
The authors would give special thanks to Mingrui Zhu (Xidian University), Zihan Ding (Princeton University), and Chenlai Qian (Southeast University) for helpful discussions and comments.

\bibliographystyle{plainnat}
\bibliography{bib}

\begin{thebibliography}{57}
\providecommand{\natexlab}[1]{#1}
\providecommand{\url}[1]{\texttt{#1}}
\expandafter\ifx\csname urlstyle\endcsname\relax
  \providecommand{\doi}[1]{doi: #1}\else
  \providecommand{\doi}{doi: \begingroup \urlstyle{rm}\Url}\fi

\bibitem[Ballester et~al.(2001)Ballester, Bertalmio, Caselles, Sapiro, and
  Verdera]{ballester2001filling}
Coloma Ballester, Marcelo Bertalmio, Vicent Caselles, Guillermo Sapiro, and
  Joan Verdera.
\newblock Filling-in by joint interpolation of vector fields and gray levels.
\newblock \emph{IEEE Transactions on Image Processing}, 10\penalty0
  (8):\penalty0 1200--1211, 2001.

\bibitem[Barnes et~al.(2009)Barnes, Shechtman, Finkelstein, and
  Goldman]{barnes2009patchmatch}
Connelly Barnes, Eli Shechtman, Adam Finkelstein, and Dan~B Goldman.
\newblock Patchmatch: A randomized correspondence algorithm for structural
  image editing.
\newblock \emph{ACM Transactions on Graphics}, 28\penalty0 (3):\penalty0 24,
  2009.

\bibitem[Bertalmio et~al.(2000)Bertalmio, Sapiro, Caselles, and
  Ballester]{bertalmio2000image}
Marcelo Bertalmio, Guillermo Sapiro, Vincent Caselles, and Coloma Ballester.
\newblock Image inpainting.
\newblock In \emph{Proceedings of the 27th annual conference on Computer
  graphics and interactive techniques}, pages 417--424, 2000.

\bibitem[Bertalmio et~al.(2003)Bertalmio, Vese, Sapiro, and
  Osher]{bertalmio2003simultaneous}
Marcelo Bertalmio, Luminita Vese, Guillermo Sapiro, and Stanley Osher.
\newblock Simultaneous structure and texture image inpainting.
\newblock \emph{IEEE Transactions on Image Processing}, 12\penalty0
  (8):\penalty0 882--889, 2003.

\bibitem[Criminisi et~al.(2004)Criminisi, P{\'e}rez, and
  Toyama]{criminisi2004region}
Antonio Criminisi, Patrick P{\'e}rez, and Kentaro Toyama.
\newblock Region filling and object removal by exemplar-based image inpainting.
\newblock \emph{IEEE Transactions on Image Processing}, 13\penalty0
  (9):\penalty0 1200--1212, 2004.

\bibitem[Deng et~al.(2018)Deng, Cheng, Xue, Zhou, and Zafeiriou]{deng2018uv}
Jiankang Deng, Shiyang Cheng, Niannan Xue, Yuxiang Zhou, and Stefanos
  Zafeiriou.
\newblock Uv-gan: Adversarial facial uv map completion for pose-invariant face
  recognition.
\newblock In \emph{CVPR}, pages 7093--7102, 2018.

\bibitem[Doersch et~al.(2012)Doersch, Singh, Gupta, Sivic, and
  Efros]{doersch2012makes}
Carl Doersch, Saurabh Singh, Abhinav Gupta, Josef Sivic, and Alexei Efros.
\newblock What makes paris look like paris?
\newblock \emph{ACM Transactions on Graphics}, 31\penalty0 (4), 2012.

\bibitem[Dolhansky and Ferrer(2018)]{dolhansky2018eye}
Brian Dolhansky and Cristian~Canton Ferrer.
\newblock Eye in-painting with exemplar generative adversarial networks.
\newblock In \emph{CVPR}, pages 7902--7911, 2018.

\bibitem[Dong et~al.(2020)Dong, Liang, Zhang, Zhang, Shen, Xie, Wu, and
  Yin]{dong2020fashion}
Haoye Dong, Xiaodan Liang, Yixuan Zhang, Xujie Zhang, Xiaohui Shen, Zhenyu Xie,
  Bowen Wu, and Jian Yin.
\newblock Fashion editing with adversarial parsing learning.
\newblock In \emph{CVPR}, pages 8120--8128, 2020.

\bibitem[Goodfellow et~al.(2014)Goodfellow, Pouget-Abadie, Mirza, Xu,
  Warde-Farley, Ozair, Courville, and Bengio]{goodfellow2014generative}
Ian Goodfellow, Jean Pouget-Abadie, Mehdi Mirza, Bing Xu, David Warde-Farley,
  Sherjil Ozair, Aaron Courville, and Yoshua Bengio.
\newblock Generative adversarial nets.
\newblock In \emph{NeurIPS}, 2014.

\bibitem[Goodfellow et~al.(2016)Goodfellow, Bengio, Courville, and
  Bengio]{goodfellow2016deep}
Ian Goodfellow, Yoshua Bengio, Aaron Courville, and Yoshua Bengio.
\newblock \emph{Deep learning}, volume~1.
\newblock MIT Press, 2016.

\bibitem[Grigorev et~al.(2019)Grigorev, Sevastopolsky, Vakhitov, and
  Lempitsky]{grigorev2019coordinate}
Artur Grigorev, Artem Sevastopolsky, Alexander Vakhitov, and Victor Lempitsky.
\newblock Coordinate-based texture inpainting for pose-guided human image
  generation.
\newblock In \emph{CVPR}, 2019.

\bibitem[Han et~al.(2019)Han, Wu, Huang, Scott, and Davis]{han2019finet}
Xintong Han, Zuxuan Wu, Weilin Huang, Matthew~R Scott, and Larry~S Davis.
\newblock Finet: Compatible and diverse fashion image inpainting.
\newblock In \emph{ICCV}, pages 4481--4491, 2019.

\bibitem[Heusel et~al.(2017)Heusel, Ramsauer, Unterthiner, Nessler, and
  Hochreiter]{heusel2017gans}
Martin Heusel, Hubert Ramsauer, Thomas Unterthiner, Bernhard Nessler, and Sepp
  Hochreiter.
\newblock Gans trained by a two time-scale update rule converge to a local nash
  equilibrium.
\newblock In \emph{NeurIPS}, 2017.

\bibitem[Iizuka et~al.(2017)Iizuka, Simo-Serra, and
  Ishikawa]{iizuka2017globally}
Satoshi Iizuka, Edgar Simo-Serra, and Hiroshi Ishikawa.
\newblock Globally and locally consistent image completion.
\newblock \emph{ACM Transactions on Graphics}, 36\penalty0 (4):\penalty0 1--14,
  2017.

\bibitem[Jordan(2003)]{jordan2003introduction}
Michael~I Jordan.
\newblock An introduction to probabilistic graphical models, 2003.

\bibitem[Karras et~al.(2017)Karras, Aila, Laine, and
  Lehtinen]{karras2017progressive}
Tero Karras, Timo Aila, Samuli Laine, and Jaakko Lehtinen.
\newblock Progressive growing of gans for improved quality, stability, and
  variation.
\newblock \emph{arXiv preprint arXiv:1710.10196}, 2017.

\bibitem[Karras et~al.(2019)Karras, Laine, and Aila]{karras2019style}
Tero Karras, Samuli Laine, and Timo Aila.
\newblock A style-based generator architecture for generative adversarial
  networks.
\newblock In \emph{CVPR}, pages 4401--4410, 2019.

\bibitem[Kingma and Ba(2015)]{kingma2015adam}
Diederik~P Kingma and Jimmy Ba.
\newblock Adam: A method for stochastic optimization.
\newblock In \emph{ICLR}, 2015.

\bibitem[Kingma and Welling(2013)]{kingma2013auto}
Diederik~P Kingma and Max Welling.
\newblock Auto-encoding variational bayes.
\newblock \emph{arXiv preprint arXiv:1312.6114}, 2013.

\bibitem[Koller and Friedman(2009)]{koller2009probabilistic}
Daphne Koller and Nir Friedman.
\newblock \emph{Probabilistic Graphical Models: Principles and Techniques}.
\newblock MIT press, 2009.

\bibitem[Levin et~al.(2003)Levin, Zomet, and Weiss]{levin2003learning}
Anat Levin, Assaf Zomet, and Yair Weiss.
\newblock Learning how to inpaint from global image statistics.
\newblock In \emph{ICCV}, pages 305--312, 2003.

\bibitem[Li et~al.(2017)Li, Liu, Yang, and Yang]{li2017generative}
Yijun Li, Sifei Liu, Jimei Yang, and Ming-Hsuan Yang.
\newblock Generative face completion.
\newblock In \emph{CVPR}, pages 3911--3919, 2017.

\bibitem[Liu et~al.(2020)Liu, Jiang, Song, Huang, and Yang]{liu2020rethinking}
Hongyu Liu, Bin Jiang, Yibing Song, Wei Huang, and Chao Yang.
\newblock Rethinking image inpainting via a mutual encoder-decoder with feature
  equalizations.
\newblock In \emph{ECCV}, pages 725--741, 2020.

\bibitem[Liu et~al.(2021)Liu, Wan, Huang, Song, Han, and Liao]{liu2021pd}
Hongyu Liu, Ziyu Wan, Wei Huang, Yibing Song, Xintong Han, and Jing Liao.
\newblock Pd-gan: Probabilistic diverse gan for image inpainting.
\newblock In \emph{CVPR}, pages 9371--9381, 2021.

\bibitem[Liu et~al.(2015)Liu, Luo, Wang, and Tang]{liu2015deep}
Ziwei Liu, Ping Luo, Xiaogang Wang, and Xiaoou Tang.
\newblock Deep learning face attributes in the wild.
\newblock In \emph{ICCV}, pages 3730--3738, 2015.

\bibitem[Mao et~al.(2017)Mao, Li, Xie, Lau, Wang, and
  Paul~Smolley]{mao2017least}
Xudong Mao, Qing Li, Haoran Xie, Raymond~YK Lau, Zhen Wang, and Stephen
  Paul~Smolley.
\newblock Least squares generative adversarial networks.
\newblock In \emph{CVPR}, pages 2794--2802, 2017.

\bibitem[McLachlan and Rathnayake(2014)]{mclachlan2014number}
Geoffrey~J McLachlan and Suren Rathnayake.
\newblock On the number of components in a gaussian mixture model.
\newblock \emph{Wiley Interdisciplinary Reviews: Data Mining and Knowledge
  Discovery}, 4\penalty0 (5):\penalty0 341--355, 2014.

\bibitem[Mohri et~al.(2018)Mohri, Rostamizadeh, and
  Talwalkar]{mohri2018foundations}
Mehryar Mohri, Afshin Rostamizadeh, and Ameet Talwalkar.
\newblock \emph{Foundations of Machine Learning}.
\newblock MIT press, 2018.

\bibitem[Nazeri et~al.(2019)Nazeri, Ng, Joseph, Qureshi, and
  Ebrahimi]{nazeri2019edgeconnect}
Kamyar Nazeri, Eric Ng, Tony Joseph, Faisal~Z Qureshi, and Mehran Ebrahimi.
\newblock Edgeconnect: Generative image inpainting with adversarial edge
  learning.
\newblock \emph{arXiv preprint arXiv:1901.00212}, 2019.

\bibitem[Parmar et~al.(2018)Parmar, Vaswani, Uszkoreit, Kaiser, Shazeer, Ku,
  and Tran]{parmar2018image}
Niki Parmar, Ashish Vaswani, Jakob Uszkoreit, Lukasz Kaiser, Noam Shazeer,
  Alexander Ku, and Dustin Tran.
\newblock Image transformer.
\newblock In \emph{ICML}, pages 4055--4064, 2018.

\bibitem[Pathak et~al.(2016)Pathak, Krahenbuhl, Donahue, Darrell, and
  Efros]{pathak2016context}
Deepak Pathak, Philipp Krahenbuhl, Jeff Donahue, Trevor Darrell, and Alexei~A
  Efros.
\newblock Context encoders: Feature learning by inpainting.
\newblock In \emph{CVPR}, pages 2536--2544, 2016.

\bibitem[Peng et~al.(2021)Peng, Liu, Xu, and Li]{peng2021generating}
Jialun Peng, Dong Liu, Songcen Xu, and Houqiang Li.
\newblock Generating diverse structure for image inpainting with hierarchical
  vq-vae.
\newblock In \emph{CVPR}, pages 10775--10784, 2021.

\bibitem[Ren et~al.(2021)Ren, Li, Ding, Pan, and Dong]{ren2021probabilistic}
Jie Ren, Yewen Li, Zihan Ding, Wei Pan, and Hao Dong.
\newblock Probabilistic mixture-of-experts for efficient deep reinforcement
  learning.
\newblock \emph{arXiv preprint arXiv:2104.09122}, 2021.

\bibitem[Ren et~al.(2015)Ren, Xu, Yan, and Sun]{ren2015shepard}
Jimmy~S Ren, Li~Xu, Qiong Yan, and Wenxiu Sun.
\newblock Shepard convolutional neural networks.
\newblock In \emph{NeurIPS}, 2015.

\bibitem[Reynolds(2009)]{reynolds2009gaussian}
Douglas~A Reynolds.
\newblock Gaussian mixture models.
\newblock \emph{Encyclopedia of Biometrics}, 741:\penalty0 659--663, 2009.

\bibitem[Russakovsky et~al.(2015)Russakovsky, Deng, Su, Krause, Satheesh, Ma,
  Huang, Karpathy, Khosla, Bernstein, et~al.]{russakovsky2015imagenet}
Olga Russakovsky, Jia Deng, Hao Su, Jonathan Krause, Sanjeev Satheesh, Sean Ma,
  Zhiheng Huang, Andrej Karpathy, Aditya Khosla, Michael Bernstein, et~al.
\newblock Imagenet large scale visual recognition challenge.
\newblock \emph{International Journal of Computer Vision}, 115\penalty0
  (3):\penalty0 211--252, 2015.

\bibitem[Van~der Maaten and Hinton(2008)]{van2008visualizing}
Laurens Van~der Maaten and Geoffrey Hinton.
\newblock Visualizing data using t-sne.
\newblock \emph{Journal of Machine Learning Research}, 9\penalty0 (11), 2008.

\bibitem[Walker et~al.(2016)Walker, Doersch, Gupta, and
  Hebert]{walker2016uncertain}
Jacob Walker, Carl Doersch, Abhinav Gupta, and Martial Hebert.
\newblock An uncertain future: Forecasting from static images using variational
  autoencoders.
\newblock In \emph{ECCV}, pages 835--851, 2016.

\bibitem[Wan et~al.(2021)Wan, Zhang, Chen, and Liao]{wan2021high}
Ziyu Wan, Jingbo Zhang, Dongdong Chen, and Jing Liao.
\newblock High-fidelity pluralistic image completion with transformers.
\newblock In \emph{ICCV}, 2021.

\bibitem[Yan et~al.(2018)Yan, Li, Li, Zuo, and Shan]{yan2018shift}
Zhaoyi Yan, Xiaoming Li, Mu~Li, Wangmeng Zuo, and Shiguang Shan.
\newblock Shift-net: Image inpainting via deep feature rearrangement.
\newblock In \emph{ECCV}, pages 1--17, 2018.

\bibitem[Yang et~al.(2019)Yang, Yu, Zheng, Yao, and Mei]{yang2019adaptive}
Shuo Yang, Wei Yu, Ying Zheng, Hongxun Yao, and Tao Mei.
\newblock Adaptive semantic-visual tree for hierarchical embeddings.
\newblock In \emph{ACM MM}, pages 2097--2105, 2019.

\bibitem[Yang et~al.(2021)Yang, Sun, Jiang, Xia, Zhang, Yuan, Wang, Luo, and
  Xu]{yang2021objects}
Shuo Yang, Peize Sun, Yi~Jiang, Xiaobo Xia, Ruiheng Zhang, Zehuan Yuan, Changhu
  Wang, Ping Luo, and Min Xu.
\newblock Objects in semantic topology.
\newblock \emph{arXiv preprint arXiv:2110.02687}, 2021.

\bibitem[Yi et~al.(2020)Yi, Tang, Azizi, Jang, and Xu]{yi2020contextual}
Zili Yi, Qiang Tang, Shekoofeh Azizi, Daesik Jang, and Zhan Xu.
\newblock Contextual residual aggregation for ultra high-resolution image
  inpainting.
\newblock In \emph{CVPR}, pages 7508--7517, 2020.

\bibitem[Yu et~al.(2018)Yu, Lin, Yang, Shen, Lu, and Huang]{yu2018generative}
Jiahui Yu, Zhe Lin, Jimei Yang, Xiaohui Shen, Xin Lu, and Thomas~S Huang.
\newblock Generative image inpainting with contextual attention.
\newblock In \emph{CVPR}, pages 5505--5514, 2018.

\bibitem[Yu et~al.(2019)Yu, Lin, Yang, Shen, Lu, and Huang]{yu2019free}
Jiahui Yu, Zhe Lin, Jimei Yang, Xiaohui Shen, Xin Lu, and Thomas~S Huang.
\newblock Free-form image inpainting with gated convolution.
\newblock In \emph{ICCV}, pages 4471--4480, 2019.

\bibitem[Zeng et~al.(2019)Zeng, Fu, Chao, and Guo]{zeng2019learning}
Yanhong Zeng, Jianlong Fu, Hongyang Chao, and Baining Guo.
\newblock Learning pyramid-context encoder network for high-quality image
  inpainting.
\newblock In \emph{CVPR}, pages 1486--1494, 2019.

\bibitem[Zeng et~al.(2021)Zeng, Lin, Lu, and Patel]{zeng2021cr}
Yu~Zeng, Zhe Lin, Huchuan Lu, and Vishal~M Patel.
\newblock Cr-fill: Generative image inpainting with auxiliary contextual
  reconstruction.
\newblock In \emph{ICCV}, pages 14164--14173, 2021.

\bibitem[Zhang et~al.(2007)Zhang, Giles, Foley, and
  Yen]{zhang2007probabilistic}
Haizheng Zhang, C~Lee Giles, Henry~C Foley, and John Yen.
\newblock Probabilistic community discovery using hierarchical latent gaussian
  mixture model.
\newblock In \emph{AAAI}, pages 663--668, 2007.

\bibitem[Zhang et~al.(2018)Zhang, Hu, Luo, Zuo, and Wang]{zhang2018semantic}
Haoran Zhang, Zhenzhen Hu, Changzhi Luo, Wangmeng Zuo, and Meng Wang.
\newblock Semantic image inpainting with progressive generative networks.
\newblock In \emph{ACM MM}, pages 1939--1947, 2018.

\bibitem[Zhao et~al.(2020)Zhao, Mo, Lin, Wang, Zuo, Chen, Xing, and
  Lu]{zhao2020uctgan}
Lei Zhao, Qihang Mo, Sihuan Lin, Zhizhong Wang, Zhiwen Zuo, Haibo Chen, Wei
  Xing, and Dongming Lu.
\newblock Uctgan: Diverse image inpainting based on unsupervised cross-space
  translation.
\newblock In \emph{CVPR}, pages 5741--5750, 2020.

\bibitem[Zheng et~al.(2019)Zheng, Cham, and Cai]{zheng2019pluralistic}
Chuanxia Zheng, Tat-Jen Cham, and Jianfei Cai.
\newblock Pluralistic image completion.
\newblock In \emph{CVPR}, pages 1438--1447, 2019.

\bibitem[Zheng et~al.(2021{\natexlab{a}})Zheng, Cham, and
  Cai]{zheng2021pluralistic}
Chuanxia Zheng, Tat-Jen Cham, and Jianfei Cai.
\newblock Pluralistic free-form image completion.
\newblock \emph{International Journal of Computer Vision}, 129\penalty0
  (10):\penalty0 2786--2805, 2021{\natexlab{a}}.

\bibitem[Zheng et~al.(2021{\natexlab{b}})Zheng, Cham, and Cai]{zheng2021tfill}
Chuanxia Zheng, Tat-Jen Cham, and Jianfei Cai.
\newblock Tfill: Image completion via a transformer-based architecture.
\newblock \emph{arXiv preprint arXiv:2104.00845}, 2021{\natexlab{b}}.

\bibitem[Zhou et~al.(2017)Zhou, Lapedriza, Khosla, Oliva, and
  Torralba]{zhou2017places}
Bolei Zhou, Agata Lapedriza, Aditya Khosla, Aude Oliva, and Antonio Torralba.
\newblock Places: A 10 million image database for scene recognition.
\newblock \emph{IEEE Transactions on Pattern Analysis and Machine
  Intelligence}, 40\penalty0 (6):\penalty0 1452--1464, 2017.

\bibitem[Zhu et~al.(2017)Zhu, Zhang, Pathak, Darrell, Efros, Wang, and
  Shechtman]{zhu2017multimodal}
Jun-Yan Zhu, Richard Zhang, Deepak Pathak, Trevor Darrell, Alexei~A Efros,
  Oliver Wang, and Eli Shechtman.
\newblock Multimodal image-to-image translation by enforcing bi-cycle
  consistency.
\newblock In \emph{NeurIPS}, pages 465--476, 2017.

\bibitem[Zong et~al.(2018)Zong, Song, Min, Cheng, Lumezanu, Cho, and
  Chen]{zong2018deep}
Bo~Zong, Qi~Song, Martin~Renqiang Min, Wei Cheng, Cristian Lumezanu, Daeki Cho,
  and Haifeng Chen.
\newblock Deep autoencoding gaussian mixture model for unsupervised anomaly
  detection.
\newblock In \emph{ICLR}, 2018.

\end{thebibliography}

\newpage
\appendix
\onecolumn

\section{The Details of Theoretical Analyses}\label{sec:A}
\subsection{The Proof of Proposition 1}\label{sec:A.1}
As we do not have the access to the underlying $\mathbf{I}_c$, we can preform the following variational approximation with the Kullback-Leibler (KL) divergence:
\begin{equation}\label{eq:variational_approximation}
\begin{aligned}
    \min&\quad\text{KL}\left[q_{\psi}(\mathbf{I}_o,\mathbf{z}_m, \mathbf{z}_c|\mathbf{I}_m)\|p_{\phi}(\mathbf{I}_o,\mathbf{z}_m, \mathbf{z}_c|\mathbf{I}_m,\mathbf{I}_c)\right]\\
    &=\int_{\mathbf{I}_o}\int_{\mathbf{z}_m}\int_{\mathbf{z}_c}q_{\psi}(\mathbf{I}_o,\mathbf{z}_m, \mathbf{z}_c|\mathbf{I}_m)\log\frac{q_{\psi}(\mathbf{I}_o,\mathbf{z}_m, \mathbf{z}_c|\mathbf{I}_m)}{p_{\phi}(\mathbf{I}_o,\mathbf{z}_m, \mathbf{z}_c|\mathbf{I}_m,\mathbf{I}_c)}\mathrm{d}\mathbf{I}_o\mathrm{d}\mathbf{z}_m\mathrm{d}\mathbf{z}_c\\
    &=\int_{\mathbf{I}_o}\int_{\mathbf{z}_m}\int_{\mathbf{z}_c}q_{\psi}(\mathbf{I}_o|\mathbf{z}_m, \mathbf{z}_c)q_{\psi}(\mathbf{z}_m, \mathbf{z}_c|\mathbf{I}_m)\log\frac{q_{\psi}(\mathbf{I}_o|\mathbf{z}_m, \mathbf{z}_c)q_{\psi}(\mathbf{z}_m, \mathbf{z}_c|\mathbf{I}_m)}{p_{\phi}(\mathbf{I}_o|\mathbf{I}_m,\mathbf{I}_c)p_{\phi}(\mathbf{z}_m, \mathbf{z}_c|\mathbf{I}_m,\mathbf{I}_c)}\mathrm{d}\mathbf{I}_o\mathrm{d}\mathbf{z}_m\mathrm{d}\mathbf{z}_c\\
    &=\int_{\mathbf{I}_o}\int_{\mathbf{z}_m}\int_{\mathbf{z}_c}q_{\psi}(\mathbf{I}_o|\mathbf{z}_m, \mathbf{z}_c)q_{\psi}(\mathbf{z}_m, \mathbf{z}_c|\mathbf{I}_m)\log\frac{q_{\psi}(\mathbf{I}_o|\mathbf{z}_m, \mathbf{z}_c)}{p_{\phi}(\mathbf{I}_o|\mathbf{I}_m,\mathbf{I}_c)}\mathrm{d}\mathbf{I}_o\mathrm{d}\mathbf{z}_m\mathrm{d}\mathbf{z}_c\\
    &+\int_{\mathbf{I}_o}\int_{\mathbf{z}_m}\int_{\mathbf{z}_c}q_{\psi}(\mathbf{I}_o|\mathbf{z}_m, \mathbf{z}_c)q_{\psi}(\mathbf{z}_m, \mathbf{z}_c|\mathbf{I}_m)\log\frac{q_{\psi}(\mathbf{z}_m, \mathbf{z}_c|\mathbf{I}_m)}{p_{\phi}(\mathbf{z}_m, \mathbf{z}_c|\mathbf{I}_m,\mathbf{I}_c)}\mathrm{d}\mathbf{I}_o\mathrm{d}\mathbf{z}_m\mathrm{d}\mathbf{z}_c\\
    &=\int_{\mathbf{z}_m}\int_{\mathbf{z}_c}q_{\psi}(\mathbf{z}_m, \mathbf{z}_c|\mathbf{I}_m)\text{KL}\left[q_{\psi}(\mathbf{I}_o|\mathbf{z}_m, \mathbf{z}_c)\|p_{\phi}(\mathbf{I}_o|\mathbf{I}_m,\mathbf{I}_c)\right]\mathrm{d}\mathbf{z}_m\mathrm{d}\mathbf{z}_c\\
    &+\int_{\mathbf{I}_o}q_{\psi}(\mathbf{I}_o|\mathbf{z}_m, \mathbf{z}_c)\text{KL}\left[q_{\psi}(\mathbf{z}_m, \mathbf{z}_c|\mathbf{I}_m)\|p_{\phi}(\mathbf{z}_m, \mathbf{z}_c|\mathbf{I}_m,\mathbf{I}_c)\right]\mathrm{d}\mathbf{I}_o\\
    &=\mathbbm{E}_{(\mathbf{z}_m,\mathbf{z}_c)\sim q_{\psi}(\mathbf{z}_m, \mathbf{z}_c|\mathbf{I}_m)}\text{KL}\left[q_{\psi}(\mathbf{I}_o|\mathbf{z}_m, \mathbf{z}_c)\|p_{\phi}(\mathbf{I}_o|\mathbf{I}_m,\mathbf{I}_c)\right]+\text{KL}\left[q_{\psi}(\mathbf{z}_m, \mathbf{z}_c|\mathbf{I}_m)\|p_{\phi}(\mathbf{z}_m, \mathbf{z}_c|\mathbf{I}_m,\mathbf{I}_c)\right]. 
\end{aligned}
\end{equation}
The second term in Eq.~(\ref{eq:variational_approximation}) is 
\begin{equation}
\begin{aligned}\label{eq:des}
&\quad\text{KL}\left[q_{\psi}(\mathbf{z}_m, \mathbf{z}_c|\mathbf{I}_m)\|p_{\phi}(\mathbf{z}_m, \mathbf{z}_c|\mathbf{I}_m,\mathbf{I}_c)\right]\\
&=\int_{\mathbf{z}_m}\int_{\mathbf{z}_c}q_{\theta}(\mathbf{z}_c|\mathbf{z}_m)q_{\psi}(\mathbf{z}_m|\mathbf{I}_m)\log\frac{q_{\theta}(\mathbf{z}_c|\mathbf{z}_m)q_{\psi}(\mathbf{z}_m|\mathbf{I}_m)}{p_{\phi}(\mathbf{z}_m, \mathbf{z}_c|\mathbf{I}_m,\mathbf{I}_c)}\mathrm{d}\mathbf{z}_m\mathrm{d}\mathbf{z}_c\\
&=\int_{\mathbf{z}_m}\int_{\mathbf{z}_c}q_{\theta}(\mathbf{z}_c|\mathbf{z}_m)q_{\psi}(\mathbf{z}_m|\mathbf{I}_m)\log\frac{q_{\theta}(\mathbf{z}_c|\mathbf{z}_m)}{p_{\phi}(\mathbf{z}_c|\mathbf{I}_c)}\mathrm{d}\mathbf{z}_m\mathrm{d}\mathbf{z}_c+q_{\theta}(\mathbf{z}_c|\mathbf{z}_m)q_{\psi}(\mathbf{z}_m|\mathbf{I}_m)\log\frac{q_{\psi}(\mathbf{z}_m|\mathbf{I}_m)}{p_{\phi}(\mathbf{z}_m|\mathbf{I}_m)}\mathrm{d}\mathbf{z}_m\mathrm{d}\mathbf{z}_c\\
&=\mathbbm{E}_{\mathbf{z}_m\sim q_{\psi}(\mathbf{z}_m|\mathbf{I}_m)}\text{KL}\left[q_{\theta}(\mathbf{z}_c|\mathbf{z}_m)\|p_{\phi}(\mathbf{z}_c|\mathbf{I}_c)\right]+\text{KL}\left[q_{\psi}(\mathbf{z}_m|\mathbf{I}_m)\|p_{\phi}(\mathbf{z}_m|\mathbf{I}_m)\right]. 
\end{aligned}
\end{equation}
Combining Eq.~(\ref{eq:variational_approximation}) and Eq.~(\ref{eq:des}), we complete the proof. 
\hfill $\blacksquare$\par
Note that the latent variables $\mathbf{z}_m$ and $\mathbf{z}_c$ are \textit{not codependent}. If $\mathbf{z}_m$ and $\mathbf{z}_c$ are codependent, the inference of $\mathbf{z}_m$ (\textit{resp.}~$\mathbf{z}_c$) \textit{must need} $\mathbf{z}_c$ (\textit{resp.}~$\mathbf{z}_m$). Then, in Eq.~(\ref{eq:des}), there should be $q_\psi(\mathbf{z}_m|\mathbf{I}_m,\mathbf{z}_c)$ and $p_\phi(\mathbf{z}_c|\mathbf{I}_c,\mathbf{z}_m)$, rather than $q_\psi(\mathbf{z}_m|\mathbf{I}_m)$ and $p_\phi(\mathbf{z}_c|\mathbf{I}_c)$. As shown in the our Figure~\ref{fig:pgm}, $\mathbf{z}_m$ and  $\mathbf{z}_c$ can be inferred without each other, but using $\mathbf{I}_m$ and $\mathbf{I}_c$. Therefore, $q_\psi(\mathbf{z}_m|\mathbf{I}_m)$ and  $p_\phi(\mathbf{z}_c|\mathbf{I}_c)$ hold in Eq.~(\ref{eq:des}).

\subsection{The Derivation of the Frequency Loss}\label{sec:A.2}
As discussed, we exploit the frequency loss \cite{ren2021probabilistic} for the Gaussian mixture model. Here, we detail The derivation of the frequency loss for our task, \textit{i.e.}, pluralistic image completion. 
Specifically, we use the KL divergence between each primitive $\mathcal{N}(\mathbf{z}_c|\bm{\mu}_i^{\hat{\mathbf{z}}_c},\bm{\Sigma}_i^{\hat{\mathbf{z}}_c})$ and  $\mathcal{N}(\mathbf{z}_c|\bm{\mu}^{\mathbf{z}_c},\bm{\Sigma}^{\mathbf{z}_c})$ as the metric of the performance:
\begin{equation}
    \normalfont{\text{KL}}\left[\mathcal{N}(\mathbf{z}_c|\bm{\mu}_i^{\hat{\mathbf{z}}_c},\bm{\Sigma}_i^{\hat{\mathbf{z}}_c})\|\mathcal{N}(\mathbf{z}_c|\bm{\mu}^{\mathbf{z}_c},\bm{\Sigma}^{\mathbf{z}_c})\right], i=1,...,k.
    \label{eq:kl_metric}
\end{equation}
Then, by comparing the KL divergence of each primitive, we can get the best primitive's index $j$. That is 
\begin{equation}
    j=\argmin_i\normalfont{\text{KL}}\left[\mathcal{N}(\mathbf{z}_c|\bm{\mu}_i^{\hat{\mathbf{z}}_c},\bm{\Sigma}_i^{\hat{\mathbf{z}}_c})\|\mathcal{N}(\mathbf{z}_c|\bm{\mu}^{\mathbf{z}_c},\bm{\Sigma}^{\mathbf{z}_c})\right].
\end{equation}
We now can give a detailed derivation for Eq.~(\ref{eq:fre}) below. Specifically, for stochastic mixture-of-experts, the gradient value of a single-instance sampling process from each primitive's performance under the metric Eq.~(\ref{eq:kl_metric}), to parameters $\theta_{\bm{\alpha}}$ that output the mixing
coefficients $\bm{\alpha}=[\alpha_1,\ldots,\alpha_k]$, can be estimated with a \textit{frequency approximate gradient}. Mathematically, we have
\begin{equation}
		\text{grad}_i = \delta_i\nabla_{\theta_i}\alpha_{i} \ \text{and} \ \delta_i = -\mathbbm{1}_i^{\text{best}} + \alpha_{i},
\end{equation}
where $\delta_i$ is the gradient of the performance metric Eq.~(\ref{eq:kl_metric}) for $\alpha_i$, $\nabla_{\theta_i}\alpha_{i}$ is the gradient of $\alpha_{i}$ for parameters $\theta_i$, and $\mathbbm{1}_i^{\text{best}}$ is the indicator function, where $\mathbbm{1}_i^{\text{best}} = 1$ if $i = j$ and $\mathbbm{1}_i^{\text{best}} = 0$ otherwise.

The accumulated frequency approximate gradient is an \textit{asymptotically unbiased estimation} of the true gradient for the sampling process from a categorical distribution $\bm{\alpha}$, with a batch of $n\rightarrow\infty$ examples. For $\delta_i$, we can detail it below:
	\begin{equation}
		\begin{aligned}
			\delta_i&= -\mathbbm{1}_i^{\text{best}} + \alpha_{i}\\
			&= \mathbbm{1}_i^{\text{best}}\alpha_{i} - \mathbbm{1}_i^{\text{best}}\alpha_{i} -\mathbbm{1}_i^{\text{best}} + \alpha_{i}\\
			&= \frac{1}{2}\mathbbm{1}_i^{\text{best}}\nabla_{\alpha_{i}}(1 - \alpha_{i})^2 + \frac{1}{2}(1 - \mathbbm{1}_i^{\text{best}})\nabla_{\alpha_{i}}\alpha_{i}^2.
		\end{aligned}
		\label{eq:lead_to_freq_loss}
	\end{equation}
Suppose a batch of examples with a number of $n$ is applied, and the true probability of the primitive $i$ to be the best primitive is $p_i$. The batch accumulated gradient will be
	\begin{equation}
		\begin{aligned}
			\overline{\text{grad}}
			=& \frac{1}{n}\sum_{l=1}^n \text{grad}_i^l=\frac{n_t}{2n}\nabla_{\alpha_{i}}(1 - \alpha_{i})^2\nabla_{\theta_i}\alpha_{i}+ \frac{(n-n_t)}{2n}\nabla_{\alpha_{i}}\alpha_{i}^2\nabla_{\theta_i}\alpha_{i}\stackrel{n \to \infty}{=}( \alpha_{i}-p_i )\nabla_{\theta_i}\alpha_{i},\\
		\end{aligned}
		\label{eq:acc_grad}
	\end{equation}

where the last formula indicates that $n_t=n p_i$ when $n \to \infty$, since the true probability can be approximated by $\frac{n_t}{n}$ in the limit case. Also, as $\nabla_{\theta_i}\alpha_{i}$ is not always equal to $0$, the gradient equals to $0$ if and only if $\alpha_{i} = p_i$ . Optimising with the above Eq.~(\ref{eq:acc_grad}) is the same as minimising the distance between $\alpha_{i}$ and $p_i$, with the optimal situation as $\alpha_{i}=p_i$ when letting the last formula of Eq.~(\ref{eq:acc_grad}) be zero. Based the above analyses, we therefore can build a frequency loss in this paper. 

The derivation is completed. 
\hfill $\blacksquare$\par

\subsection{The Proof of Proposition 2}\label{sec:A.3}

We employ the \textit{mixture of Gaussian} for the \textit{dynamic} model $q_{\theta}(\mathbf{z}_c|\mathbf{z}_m)$, which is more flexible than a unimodal and can diversify the output. More formally, we have 
\begin{equation}
\begin{aligned}
q_{\theta}(\mathbf{z}_c|\mathbf{z}_m)=\sum_{i=1}^k \alpha_i\mathcal{N}(\mathbf{z}_c|\bm{\mu}_i^{\hat{\mathbf{z}}_c},\bm{\Sigma}_i^{\hat{\mathbf{z}}_c}),
\end{aligned}
\end{equation}
where $k$ denotes the number of primitives, and $\alpha_i$ denotes the weight of the $i$-th primitive. Then the KL divergence between $q_{\theta}(\mathbf{z}_c|\mathbf{z}_m)$ and $p_{\phi}(\mathbf{z}_c|\mathbf{I}_c)$ becomes 
\begin{equation}\label{eq:kl_gaussian}
\begin{aligned}
&\quad\text{KL}\left[q_{\theta}(\mathbf{z}_c|\mathbf{z}_m)\|p_{\phi}(\mathbf{z}_c|\mathbf{I}_c)\right]\\
&=\int_{\mathbf{z}_c}q_{\theta}(\mathbf{z}_c|\mathbf{z}_m)\log\frac{q_{\theta}(\mathbf{z}_c|\mathbf{z}_m)}{p_{\phi}(\mathbf{z}_c|\mathbf{I}_c)}\mathrm{d}\mathbf{z}_c\\
&=\int_{\mathbf{z}_c}\sum_{i=1}^k \alpha_i\mathcal{N}(\mathbf{z}_c|\bm{\mu}_i^{\hat{\mathbf{z}}_c},\bm{\Sigma}_j^{\hat{\mathbf{z}}_c})\log\frac{\sum_{i=1}^k \alpha_i\mathcal{N}(\mathbf{z}_c|\bm{\mu}_i^{\hat{\mathbf{z}}_c},\bm{\Sigma}_i^{\hat{\mathbf{z}}_c})}{\mathcal{N}(\mathbf{z}_c|\bm{\mu}^{\mathbf{z}_c},\bm{\Sigma}^{\mathbf{z}_c})}\mathrm{d}\mathbf{z}_c.
\end{aligned}
\end{equation}
We further approximate Eq.~(\ref{eq:kl_gaussian}) by the back-propogate-max-operation with the frequency loss. We will have 
\begin{equation}\label{eq:max_approx}
\begin{aligned}
&\quad\text{KL}\left[q_{\theta}(\mathbf{z}_c|\mathbf{z}_m)\|p_{\phi}(\mathbf{z}_c|\mathbf{I}_c)\right]\approx\int_{\mathbf{z}_c}\mathcal{N}(\mathbf{z}_c|\bm{\mu}_j^{\hat{\mathbf{z}}_c},\bm{\Sigma}_j^{\hat{\mathbf{z}}_c})\log\frac{\mathcal{N}(\mathbf{z}_c|\bm{\mu}_j^{\hat{\mathbf{z}}_c},\bm{\Sigma}_j^{\hat{\mathbf{z}}_c})}{\mathcal{N}(\mathbf{z}_c|\bm{\mu}^{\mathbf{z}_c},\bm{\Sigma}^{\mathbf{z}_c})}\mathrm{d}\mathbf{z}_c, 
\end{aligned}
\end{equation}
where $j=\argmin_i\text{KL}\left[\mathcal{N}(\mathbf{z}_c|\bm{\mu}_i^{\hat{\mathbf{z}}_c},\bm{\Sigma}_i^{\hat{\mathbf{z}}_c})\|\mathcal{N}(\mathbf{z}_c|\bm{\mu}^{\mathbf{z}_c},\bm{\Sigma}^{\mathbf{z}_c})\right]$. We then have 
\begin{equation}
\begin{aligned}
&\quad\int_{\mathbf{z}_c}\mathcal{N}(\mathbf{z}_c|\bm{\mu}_j^{\hat{\mathbf{z}}_c},\bm{\Sigma}_j^{\hat{\mathbf{z}}_c})\log\frac{\mathcal{N}(\mathbf{z}_c|\bm{\mu}_j^{\hat{\mathbf{z}}_c},\bm{\Sigma}_j^{\hat{\mathbf{z}}_c})}{\mathcal{N}(\mathbf{z}_c|\bm{\mu}^{\mathbf{z}_c},\bm{\Sigma}^{\mathbf{z}_c})}\mathrm{d}\mathbf{z}_c\\
&=\int_{\mathbf{z}_c}\mathcal{N}(\mathbf{z}_c|\bm{\mu}_j^{\hat{\mathbf{z}}_c},\bm{\Sigma}_j^{\hat{\mathbf{z}}_c})\log\frac{(2\pi)^{-\frac{d}{2}}{|\bm{\Sigma}_j^{\hat{\mathbf{z}}_c}|}^{-\frac{1}{2}}\exp\left[-\frac{1}{2}(\mathbf{z}_c-\bm{\mu}_j^{\hat{\mathbf{z}}_c})^{\top}{(\bm{\Sigma}_j^{\hat{\mathbf{z}}_c})}^{-1}(\mathbf{z}_c-\bm{\mu}_j^{\hat{\mathbf{z}}_c})\right]}{(2\pi)^{-\frac{d}{2}}{|\bm{\Sigma}^{\mathbf{z}_c}|}^{-\frac{1}{2}}\exp\left[-\frac{1}{2}(\mathbf{z}_c-\bm{\mu}^{\mathbf{z}_c})^{\top}{(\bm{\Sigma}^{\mathbf{z}_c})}^{-1}(\mathbf{z}_c-\bm{\mu}^{\mathbf{z}_c})\right]}\mathrm{d}\mathbf{z}_c\\
&=\int_{\mathbf{z}_c}\mathcal{N}(\mathbf{z}_c|\bm{\mu}_j^{\hat{\mathbf{z}}_c},\bm{\Sigma}_j^{\hat{\mathbf{z}}_c})\log\frac{{|\bm{\Sigma}_j^{\hat{\mathbf{z}}_c}|}^{-\frac{1}{2}}\exp\left[-\frac{1}{2}(\mathbf{z}_c-\bm{\mu}_j^{\hat{\mathbf{z}}_c})^{\top}{(\bm{\Sigma}_j^{\hat{\mathbf{z}}_c})}^{-1}(\mathbf{z}_c-\bm{\mu}_j^{\hat{\mathbf{z}}_c})\right]}{{|\bm{\Sigma}^{\mathbf{z}_c}|}^{-\frac{1}{2}}\exp\left[-\frac{1}{2}(\mathbf{z}_c-\bm{\mu}^{\mathbf{z}_c})^{\top}{(\bm{\Sigma}^{\mathbf{z}_c})}^{-1}(\mathbf{z}_c-\bm{\mu}^{\mathbf{z}_c})\right]}\mathrm{d}\mathbf{z}_c\\
&=\int_{\mathbf{z}_c}\mathcal{N}(\mathbf{z}_c|\bm{\mu}_j^{\hat{\mathbf{z}}_c},\bm{\Sigma}_j^{\hat{\mathbf{z}}_c})\{-\frac{1}{2}\log\frac{|\bm{\Sigma}_j^{\hat{\mathbf{z}}_c}|}{|\bm{\Sigma}^{\mathbf{z}_c}|}+\frac{1}{2}[(\mathbf{z}_c-\bm{\mu}^{\mathbf{z}_c})^{\top}{(\bm{\Sigma}^{\mathbf{z}_c})}^{-1}(\mathbf{z}_c-\bm{\mu}^{\mathbf{z}_c})-(\mathbf{z}_c-\bm{\mu}_j^{\hat{\mathbf{z}}_c})^{\top}{(\bm{\Sigma}_j^{\hat{\mathbf{z}}_c})}^{-1}(\mathbf{z}_c-\bm{\mu}_j^{\hat{\mathbf{z}}_c})]\}\mathrm{d}\mathbf{z}_c\\
&= -\frac{1}{2}\log\frac{|\bm{\Sigma}_j^{\hat{\mathbf{z}}_c}|}{|\bm{\Sigma}^{\mathbf{z}_c}|} + \frac{1}{2}\int_{\mathbf{z}_c}\mathcal{N}(\mathbf{z}_c|\bm{\mu}_j^{\hat{\mathbf{z}}_c},\bm{\Sigma}_j^{\hat{\mathbf{z}}_c})\Big[{\mathbf{z}_c}^{\top}{(\bm{\Sigma}^{\mathbf{z}_c})}^{-1}\mathbf{z}_c -{\mathbf{z}_c}^{\top}{(\bm{\Sigma}^{\mathbf{z}_c})}^{-1}\bm{\mu}^{\mathbf{z}_c}
-{\bm{\mu}^{\mathbf{z}_c}}^{\top}{(\bm{\Sigma}^{\mathbf{z}_c})}^{-1}\mathbf{z}_c
+ {\bm{\mu}^{\mathbf{z}_c}}^\top{(\bm{\Sigma}^{\mathbf{z}_c})}^{-1}\bm{\mu}^{\mathbf{z}_c}\\
&- ({\mathbf{z}_c}^{\top}{(\bm{\Sigma}_j^{\hat{\mathbf{z}}_c})}^{-1}\mathbf{z}_c
-{\mathbf{z}_c}^{\top}{(\bm{\Sigma}_j^{\hat{\mathbf{z}}_c})}^{-1}\bm{\mu}_j^{\hat{\mathbf{z}}_c}
-{\bm{\mu}_j^{\hat{\mathbf{z}}_c}}^{\top}{(\bm{\Sigma}_j^{\hat{\mathbf{z}}_c})}^{-1}\mathbf{z}_c
+ {\bm{\mu}_j^{\hat{\mathbf{z}}_c}}^\top{(\bm{\Sigma}_j^{\hat{\mathbf{z}}_c})}^{-1}\bm{\mu}_j^{\hat{\mathbf{z}}_c})\Big]\mathrm{d}\mathbf{z}_c.
\label{eq:elbo_without_assumption}
\end{aligned}
\end{equation}
Suppose the covariance is a diagonal matrix \cite{kingma2013auto}, we have
\begin{equation}
    \int_{\mathbf{z}_c}\mathcal{N}(\mathbf{z}_c|\bm{\mu}_j^{\hat{\mathbf{z}}_c},\bm{\Sigma}_j^{\hat{\mathbf{z}}_c}){\mathbf{z}_c}^{\top}\mathbf{z}_c\mathrm{d}\mathbf{z}_c = \text{tr}(\bm{\Sigma}_j^{\hat{\mathbf{z}}_c}) + {\bm{\mu}_j^{\hat{\mathbf{z}}_c}}^{\top}\bm{\mu}_j^{\hat{\mathbf{z}}_c},
\end{equation}
where $\text{tr}(\cdot)$ denotes the trace of a matrix. Then, Eq.~(\ref{eq:elbo_without_assumption}) would be:
\begin{equation}
\begin{aligned}
&\int_{\mathbf{z}_c}\mathcal{N}(\mathbf{z}_c|\bm{\mu}_j^{\hat{\mathbf{z}}_c},\bm{\Sigma}_j^{\hat{\mathbf{z}}_c})\log\frac{\mathcal{N}(\mathbf{z}_c|\bm{\mu}_j^{\hat{\mathbf{z}}_c},\bm{\Sigma}_j^{\hat{\mathbf{z}}_c})}{\mathcal{N}(\mathbf{z}_c|\bm{\mu}^{\mathbf{z}_c},\bm{\Sigma}^{\mathbf{z}_c})}\mathrm{d}\mathbf{z}_c\\
&= -\frac{1}{2}\log\frac{|\bm{\Sigma}_j^{\hat{\mathbf{z}}_c}|}{|\bm{\Sigma}^{\mathbf{z}_c}|}
-\frac{1}{2}\Big[\text{tr}\left({(\bm{\Sigma}^{\mathbf{z}_c})}^{-1}\bm{\Sigma}_j^{\hat{\mathbf{z}}_c}\right)
+{\bm{\mu}_j^{\hat{\mathbf{z}}_c}}^{\top}{(\bm{\Sigma}^{\mathbf{z}_c})}^{-1}\bm{\mu}_j^{\hat{\mathbf{z}}_c}
-{\bm{\mu}_j^{\hat{\mathbf{z}}_c}}^\top{(\bm{\Sigma}^{\mathbf{z}_c})}^{-1}\bm{\mu}^{\mathbf{z}_c}
-{\bm{\mu}^{\mathbf{z}_c}}^\top{(\bm{\Sigma}^{\mathbf{z}_c})}^{-1}\bm{\mu}_j^{\hat{\mathbf{z}}_c}\\
&+{\bm{\mu}^{\mathbf{z}_c}}^\top{(\bm{\Sigma}^{\mathbf{z}_c})}^{-1}\bm{\mu}^{\mathbf{z}_c}
+\text{tr}\left({(\bm{\Sigma}_j^{\hat{\mathbf{z}}_c})}^{-1}\bm{\Sigma}_j^{\hat{\mathbf{z}}_c}\right)
+{\bm{\mu}_j^{\hat{\mathbf{z}}_c}}^{\top}{(\bm{\Sigma}_j^{\hat{\mathbf{z}}_c})}^{-1}\bm{\mu}_j^{\hat{\mathbf{z}}_c}
-{\bm{\mu}_j^{\hat{\mathbf{z}}_c}}^\top{(\bm{\Sigma}_j^{\hat{\mathbf{z}}_c})}^{-1}\bm{\mu}_j^{\hat{\mathbf{z}}_c}
-{\bm{\mu}_j^{\hat{\mathbf{z}}_c}}^\top{(\bm{\Sigma}_j^{\hat{\mathbf{z}}_c})}^{-1}\bm{\mu}_j^{\hat{\mathbf{z}}_c}\\
&+{\bm{\mu}_j^{\hat{\mathbf{z}}_c}}^\top{(\bm{\Sigma}_j^{\hat{\mathbf{z}}_c})}^{-1}\bm{\mu}_j^{\hat{\mathbf{z}}_c}\Big]\\
&= -\frac{1}{2}\log\frac{|\bm{\Sigma}_j^{\hat{\mathbf{z}}_c}|}{|\bm{\Sigma}^{\mathbf{z}_c}|}
+ \frac{1}{2}\text{tr}\left({(\bm{\Sigma}^{\mathbf{z}_c})}^{-1}\bm{\Sigma}_j^{\hat{\mathbf{z}}_c}\right)
- \frac{1}{2}(\bm{\mu}_j^{\hat{\mathbf{z}}_c} - \bm{\mu}^{\mathbf{z}_c})^\top{(\bm{\Sigma}^{\mathbf{z}_c})}^{-1}(\bm{\mu}_j^{\hat{\mathbf{z}}_c} - \bm{\mu}^{\mathbf{z}_c}).
\end{aligned}
\end{equation}
The proof is completed. 
\hfill $\blacksquare$\par

\subsection{The Fully Optimization Expression of Proposition 1}\label{sec:A.4}
As discussed in the main paper, we have 
\begin{equation}\label{eq:analytical_a}
\begin{aligned}
    &L_{\normalfont{{\text{P}}}}=\normalfont{\text{KL}}\left[q_{\psi}(\mathbf{I}_o,\mathbf{z}_m, \mathbf{z}_c|\mathbf{I}_m)\|p_{\phi}(\mathbf{I}_o,\mathbf{z}_m, \mathbf{z}_c|\mathbf{I}_m,\mathbf{I}_c)\right]\\
    &=\underbrace{\mathbbm{E}_{(\mathbf{z}_m,\mathbf{z}_c)\sim q_{\psi}(\mathbf{z}_m, \mathbf{z}_c|\mathbf{I}_m)}\normalfont{\text{KL}}\left[q_{\psi}(\mathbf{I}_o|\mathbf{z}_m, \mathbf{z}_c)\|p_{\phi}(\mathbf{I}_o|\mathbf{I}_m,\mathbf{I}_c)\right]}_{\textcircled{a}}\\
    &+\underbrace{\mathbbm{E}_{\mathbf{z}_m\sim q_{\psi}(\mathbf{z}_m|\mathbf{I}_m)}\normalfont{\text{KL}}\left[q_{\theta}(\mathbf{z}_c|\mathbf{z}_m)\|p_{\phi}(\mathbf{z}_c|\mathbf{I}_c)\right]}_{\textcircled{b}}\\
    &+\underbrace{\normalfont{\text{KL}}\left[q_{\psi}(\mathbf{z}_m|\mathbf{I}_m)\|p_{\phi}(\mathbf{z}_m|\mathbf{I}_m)\right]}_{\textcircled{c}}.
\end{aligned}
\end{equation}

As Eq.~(\ref{eq:analytical_a}) shows,  the item \textcircled{c} may be hard to implement, because we cannot know the real $p_\phi(\mathbf{z}_m|\mathbf{I}_m)$, which is used to restrain $q_\psi(\mathbf{z}_m|\mathbf{I}_m)$ outputted by a neural network. 

However, inspired by \cite{kingma2013auto}, we can employ a variational evidence lower bound (ELBO) to implement the item \textcircled{c}. Firstly, we assume that there exists a ground truth model $p(\mathbf{I}_m)$, which is a constant given $\mathbf{I}_m$ and does not depend on $\mathbf{z}_m$. Then, we can derive the relationship between $p(\mathbf{I}_m)$ and \textcircled{c}:
\begin{equation}
\begin{aligned}
        \log p(\mathbf{I}_m) &= \mathbbm{E}_{\mathbf{z}_m \sim q_\psi(\mathbf{z}_m|\mathbf{I}_m)}[\log p(\mathbf{I}_m)]\\
        &=  \mathbbm{E}_{\mathbf{z}_m \sim q_\psi(\mathbf{z}_m|\mathbf{I}_m)} [\log \frac{p(\mathbf{I}_m|\mathbf{z}_m)p(\mathbf{z}_m)}{p_\phi(\mathbf{z}_m|\mathbf{I}_m)}] \\
        &= \mathbbm{E}_{\mathbf{z}_m \sim q_\psi(\mathbf{z}_m|\mathbf{I}_m)}[\log \frac{p(\mathbf{I}_m|\mathbf{z}_m)p(\mathbf{z}_m)}{p_\phi(\mathbf{z}_m|\mathbf{I}_m)}\frac{q_\psi(\mathbf{z}_m|\mathbf{I}_m)}{q_\psi(\mathbf{z}_m|\mathbf{I}_m)}]\\
        &=\mathbbm{E}_{\mathbf{z}_m}[\log p(\mathbf{I}_m|\mathbf{z}_m)] - \mathbbm{E}_{\mathbf{z}_m}[\log \frac{q_\psi(\mathbf{z}_m|\mathbf{I}_m)}{p(\mathbf{z}_m)}] + \mathbbm{E}_{\mathbf{z}_m}[\log \frac{q_\psi(\mathbf{z}_m|\mathbf{I}_m)}{p_\phi(\mathbf{z}_m|\mathbf{I}_m)}]\\
        &=\mathbbm{E}_{\mathbf{z}_m}[\log p(\mathbf{I}_m|\mathbf{z}_m)] - \normalfont{\text{KL}}[q_\psi(\mathbf{z}_m|\mathbf{I}_m)||p(\mathbf{z}_m)] + \normalfont{\text{KL}}[q_\psi(\mathbf{z}_m|\mathbf{I}_m)||p_\phi(\mathbf{z}_m|\mathbf{I}_m)]\\
        &=\underbrace{\mathbbm{E}_{\mathbf{z}_m}[\log p(\mathbf{I}_m|\mathbf{z}_m)] - \normalfont{\text{KL}}[q_\psi(\mathbf{z}_m|\mathbf{I}_m)||p(\mathbf{z}_m)]}_{\operatorname{ELBO}} +  \textcircled{c}         
\end{aligned}
\end{equation}
Thus,  \textcircled{c} is equal to $\log p(\mathbf{I}_m) - \operatorname{ELBO}$. Since $\log p(\mathbf{I}_m)$ is a constant, minimizing \textcircled{c} is equal to maximize the $\operatorname{ELBO}$. 

Accordingly, the fully optimization expression of Theorem 1 is below:
\begin{equation}
\begin{aligned}
    &L_{\normalfont{{\text{P}}}}=\normalfont{\text{KL}}\left[q_{\psi}(\mathbf{I}_o,\mathbf{z}_m, \mathbf{z}_c|\mathbf{I}_m)\|p_{\phi}(\mathbf{I}_o,\mathbf{z}_m, \mathbf{z}_c|\mathbf{I}_m,\mathbf{I}_c)\right]\\
    &=\underbrace{\mathbbm{E}_{(\mathbf{z}_m,\mathbf{z}_c)\sim q_{\psi}(\mathbf{z}_m, \mathbf{z}_c|\mathbf{I}_m)}\normalfont{\text{KL}}\left[q_{\psi}(\mathbf{I}_o|\mathbf{z}_m, \mathbf{z}_c)\|p_{\phi}(\mathbf{I}_o|\mathbf{I}_m,\mathbf{I}_c)\right]}_{\textcircled{a}}\\
    &+\underbrace{\mathbbm{E}_{\mathbf{z}_m\sim q_{\psi}(\mathbf{z}_m|\mathbf{I}_m)}\normalfont{\text{KL}}\left[q_{\theta}(\mathbf{z}_c|\mathbf{z}_m)\|p_{\phi}(\mathbf{z}_c|\mathbf{I}_c)\right]}_{\textcircled{b}}\\
    &-\underbrace{\mathbbm{E}_{\mathbf{z}_m \sim q_\psi(\mathbf{z}_m|\mathbf{I}_m)}[\log p(\mathbf{I}_m|\mathbf{z}_m)] +\normalfont{\text{KL}}[q_\psi(\mathbf{z}_m|\mathbf{I}_m)||p(\mathbf{z}_m)]}_{\textcircled{c}},
\end{aligned}
\end{equation}
\hfill $\blacksquare$\par
\section{Advance of PICME Over PIC}\label{sec:B}
We also summarize the differences between PIC and our method as follows:

(1) Our method has \textit{better interpretability} in pluralistic image completion than PIC, which is highlighted in this paper.

(2) GMM is used in our method, which better promotes diversity than PIC. More importantly, the inherent parameters for diversity are  \textit{task-related}, rather than \textit{task-agnostic} in PIC. 

(3) Our method uses KL divergence for the whole objective. We decompose the divergence and then minimize different decomposition terms, which supports our theory. While, PIC only uses KL divergence as a single loss in its objective. 

(4) Our method exploits the same CNN backbone as PIC. In all quantitative comparisons, our method outperforms PIC. 
\section{Supplementary Experimental Results on Diversity Analyses}\label{sec:C}
\begin{figure*}[!h]
    \centering
    \includegraphics[width=0.48\textwidth]{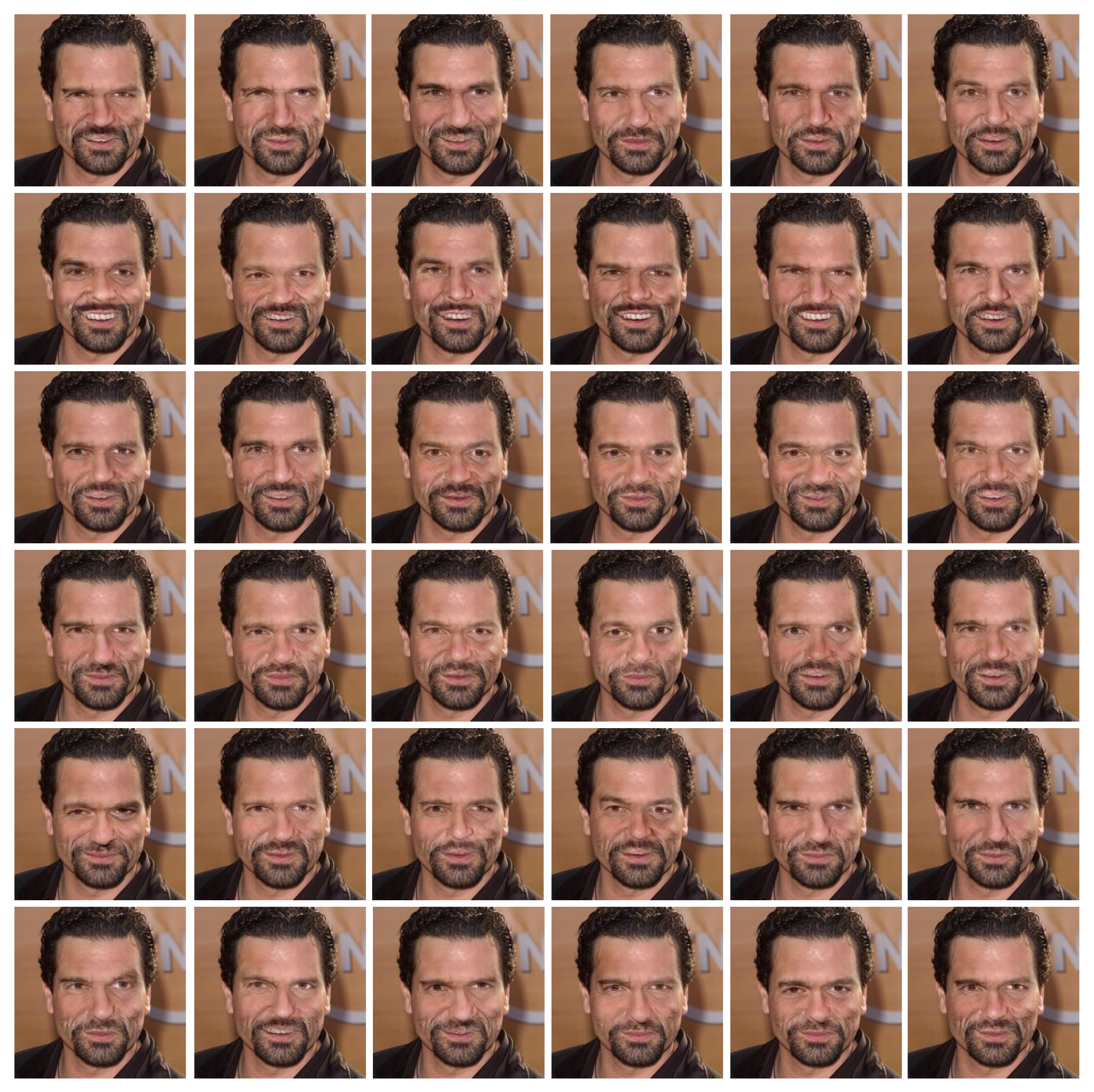}
    \caption{Pluralistic image completion results of our method by sampling from $k$ primitives ($k=6$). With each primitive, six diverse images are generated.}
    \label{fig:gmmsam}
\end{figure*}

\begin{figure*}[!h]
    \centering
    \includegraphics[width=0.48\textwidth]{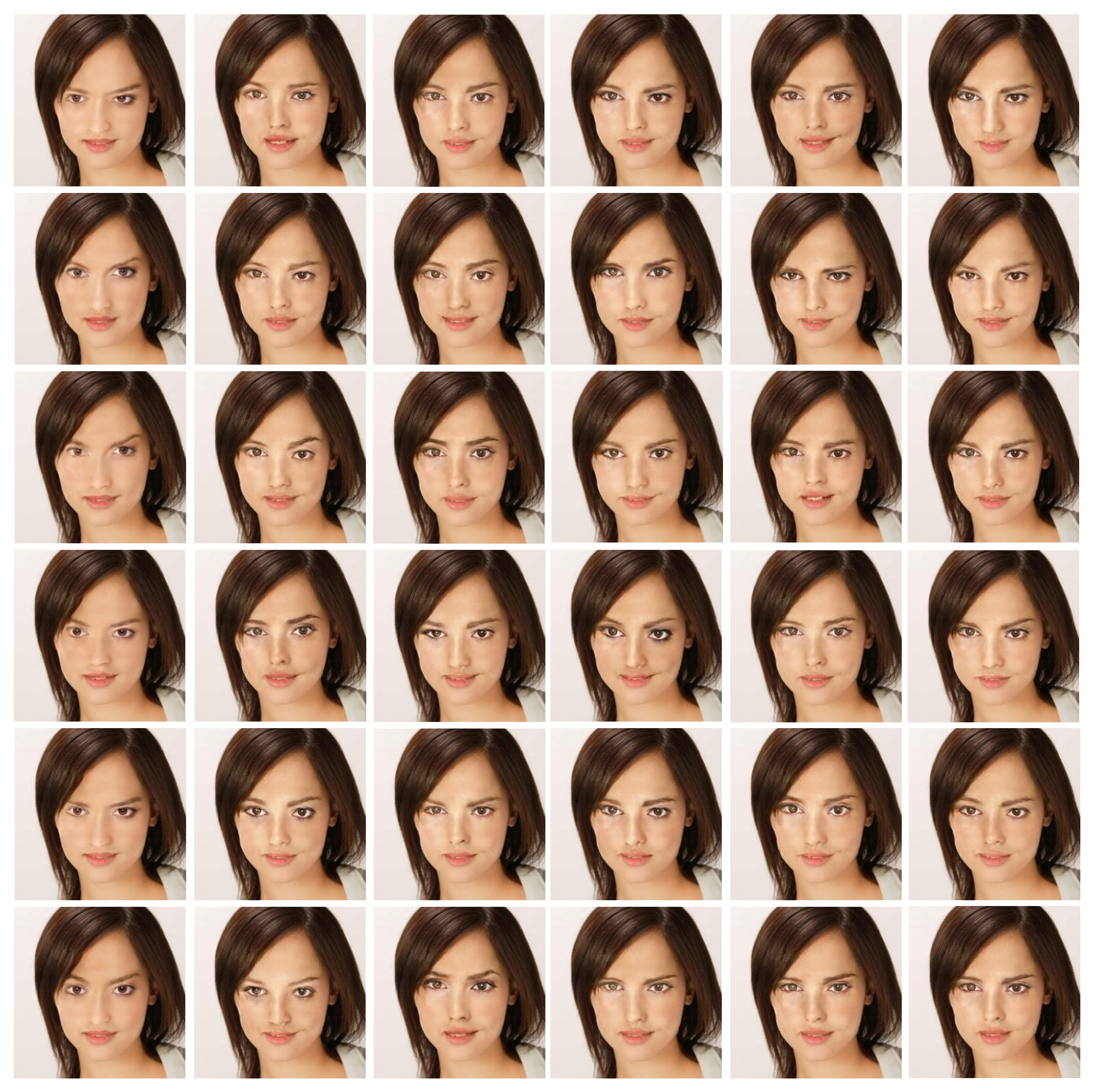}
    \caption{Pluralistic image completion results of our method by sampling from $k$ primitives ($k=6$). With each primitive, six diverse images are generated.}
    \label{fig:gmmsam2}
\end{figure*}

In the main paper, we state that our method is not limited to only generate $k$ image completion results. Instead, we can sample from the $k$ primitives of GMM to generate a different number of images. The results are provided in Figure~{\ref{fig:gmmsam}} and {\ref{fig:gmmsam2}}. The images in the $i$-th row of Figure~{\ref{fig:gmmsam}} and {\ref{fig:gmmsam2}} are obtained by sampling from the $i$-th primitive of GMM. As can be seen, the images in the same rows are less diverse than the images in the different rows, since one primitive has a limited capacity. Multiple primitives can better meet the diversity needs. We are able to sample from different primitives to achieve the better diversity in pluralistic image completion. 

\vspace{-10pt}
\clearpage

\section{Supplementary Pluralistic Completion Result Comparisons}\label{sec:D} 
\begin{figure*}[!h]
    \centering
    \includegraphics[width=\textwidth]{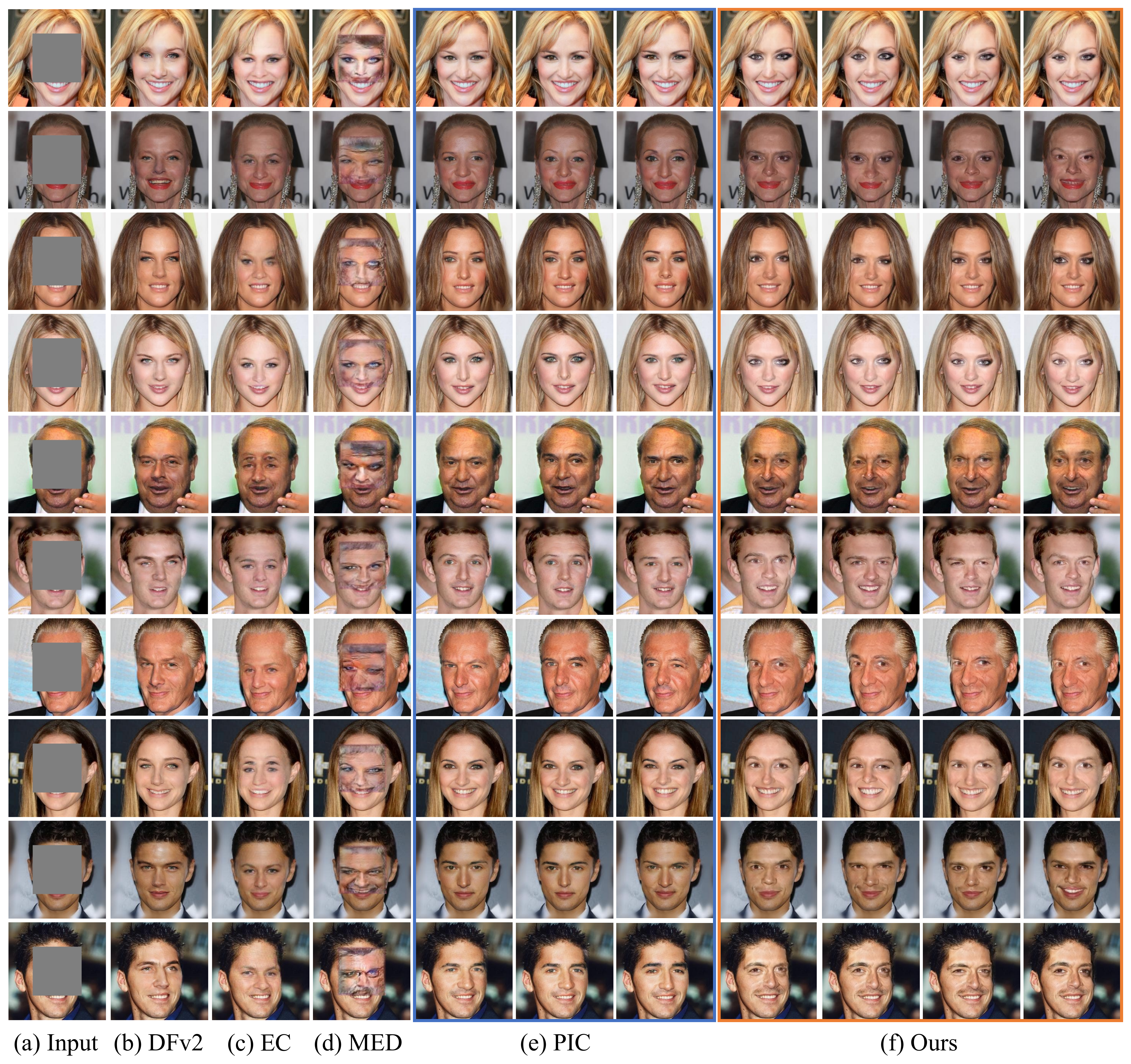}
    \caption{Pluralistic image completion results comparison with baselines. The original images come from CelebA-HQ. Best viewed by zooming in.}
    \label{fig:supp_celeba_1}
\end{figure*}

\begin{figure*}[!h]
    \centering
    \includegraphics[width=\textwidth]{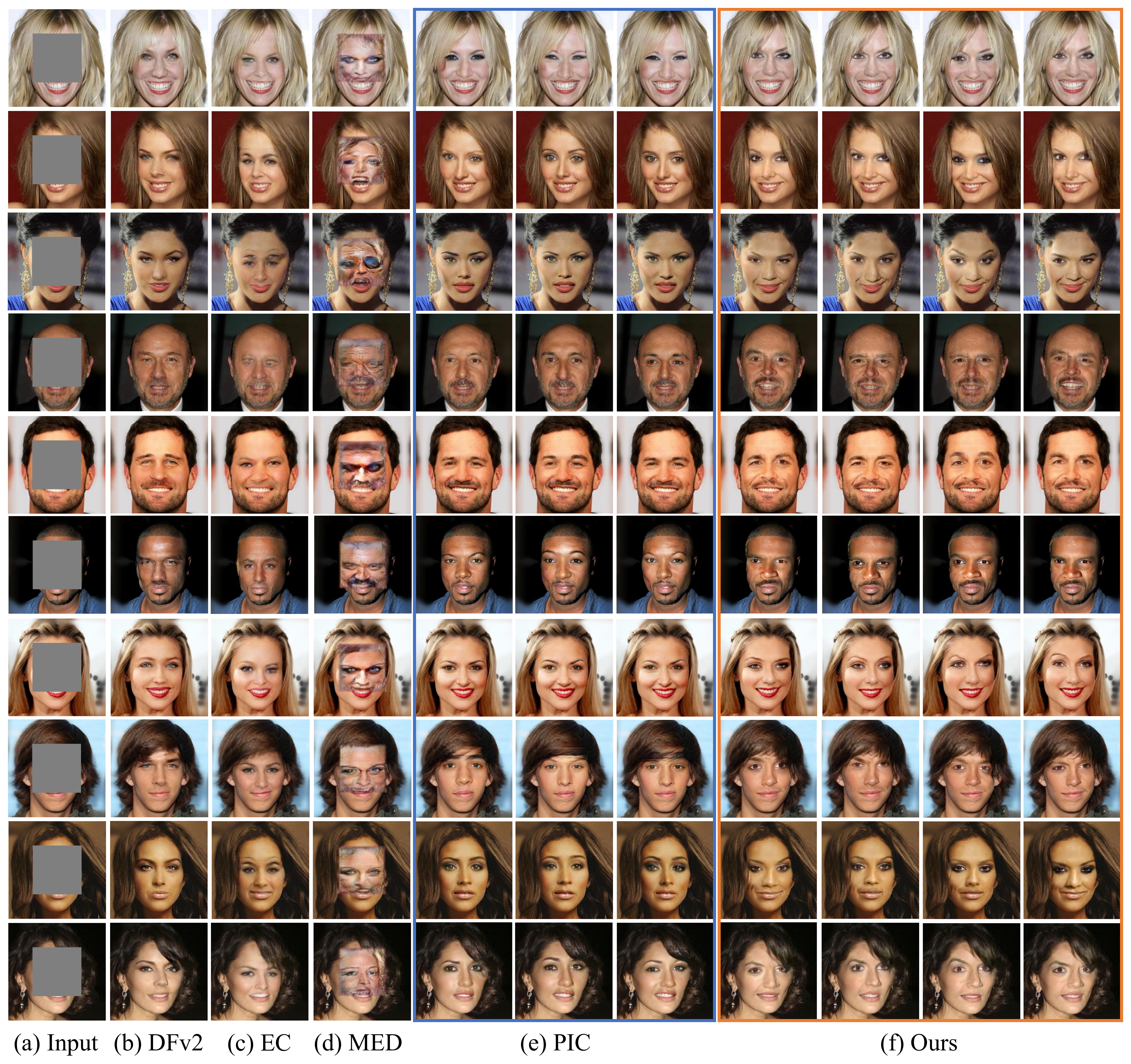}
    \caption{Pluralistic image completion results comparison with baselines. The original images come from CelebA-HQ. Best viewed by zooming in.}
    \label{fig:supp_celeba_2}
\end{figure*}

\begin{figure*}[!h]
    \centering
    \includegraphics[width=\textwidth]{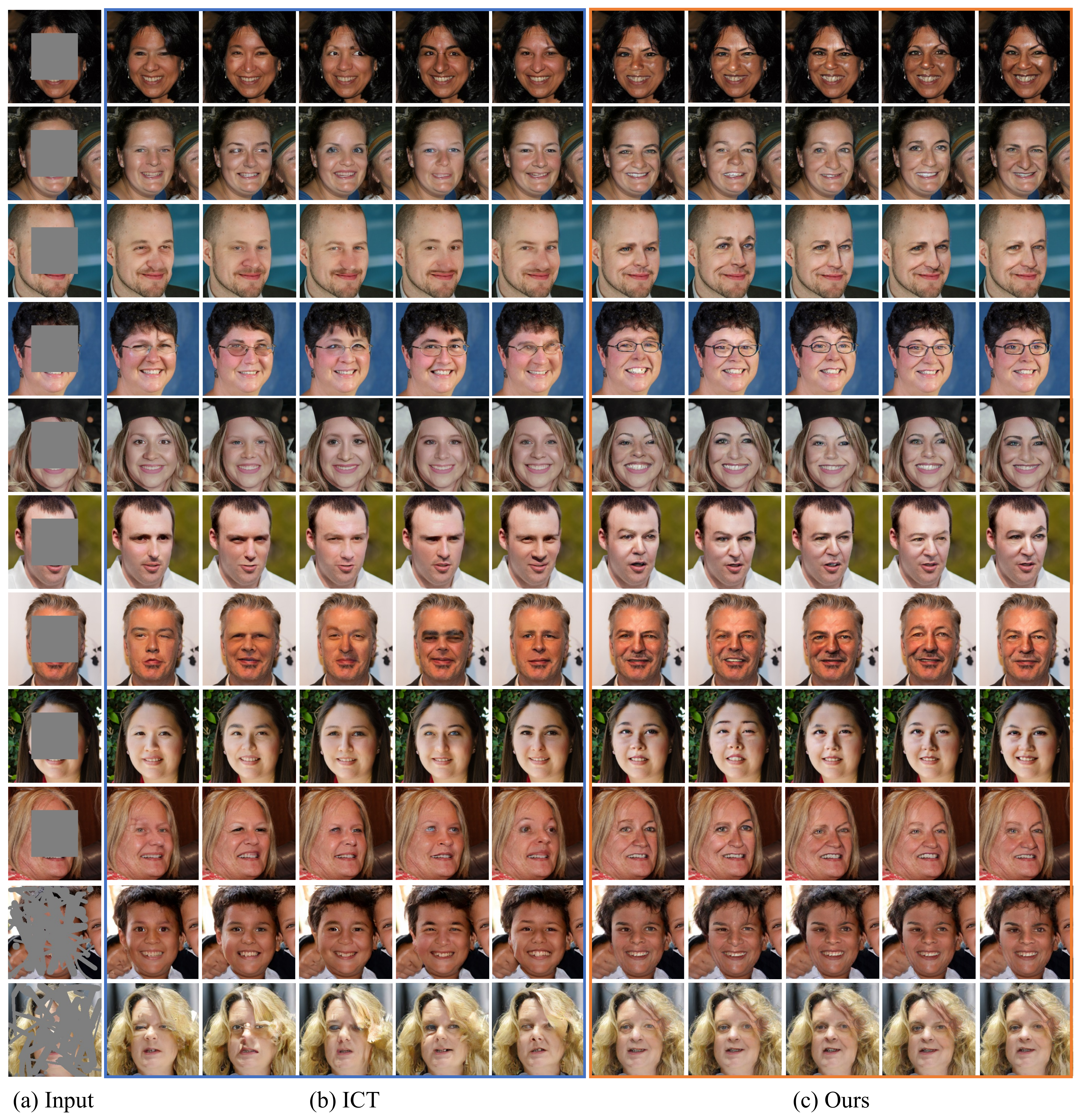}
    \caption{Pluralistic image completion results comparison with baselines. The original images come from FFHQ. Best viewed by zooming in.}
    \label{fig:supp_ffhq}
\end{figure*}

\begin{figure*}[!h]
    \centering
    \includegraphics[width=\textwidth]{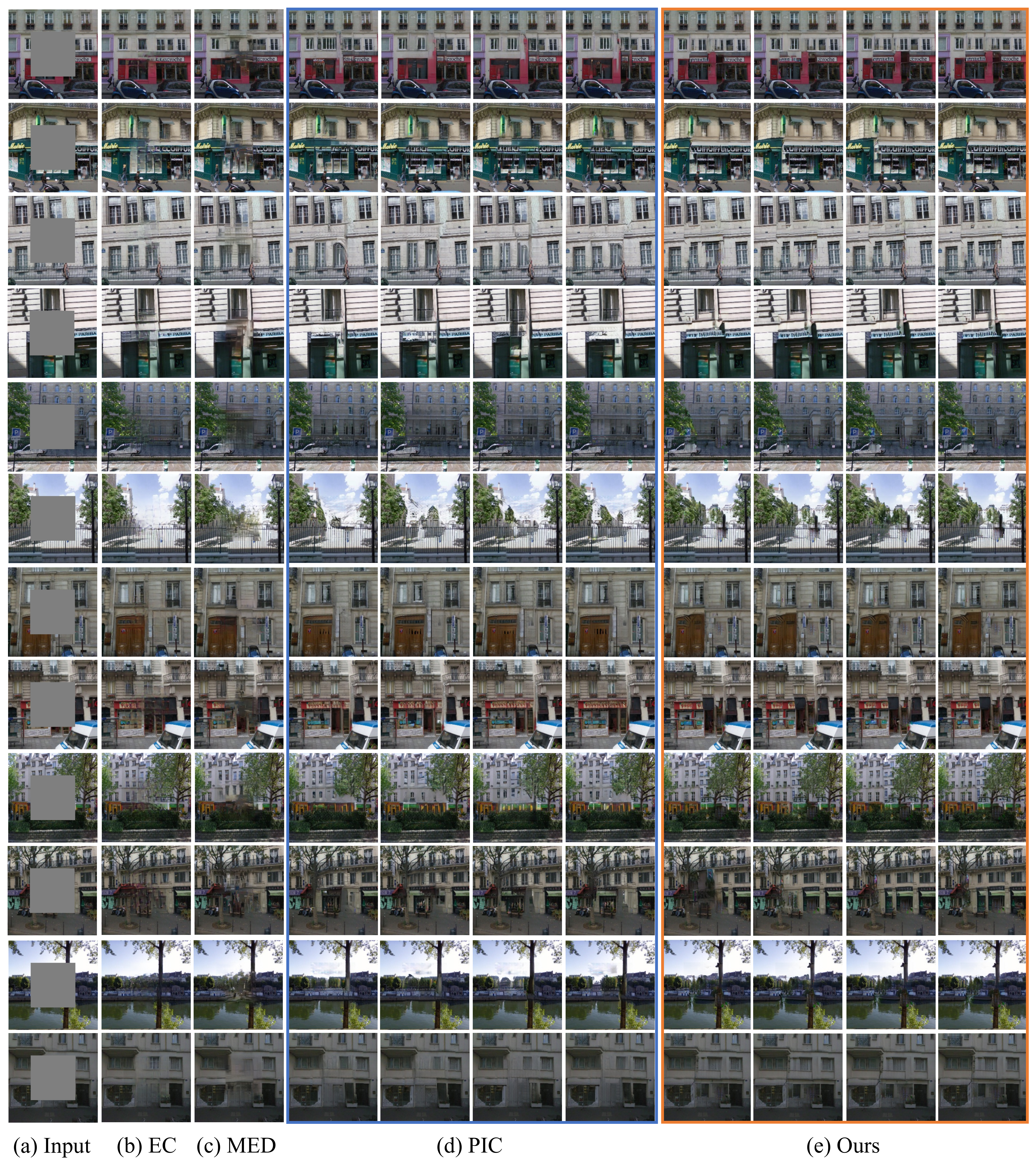}
    \caption{Pluralistic image completion results comparison with baselines. The original images come from Paris StreetView. Best viewed by zooming in.}
    \label{fig:supp_paris}
\end{figure*}

\begin{figure*}[!h]
    \centering
    \includegraphics[width=0.95\textwidth]{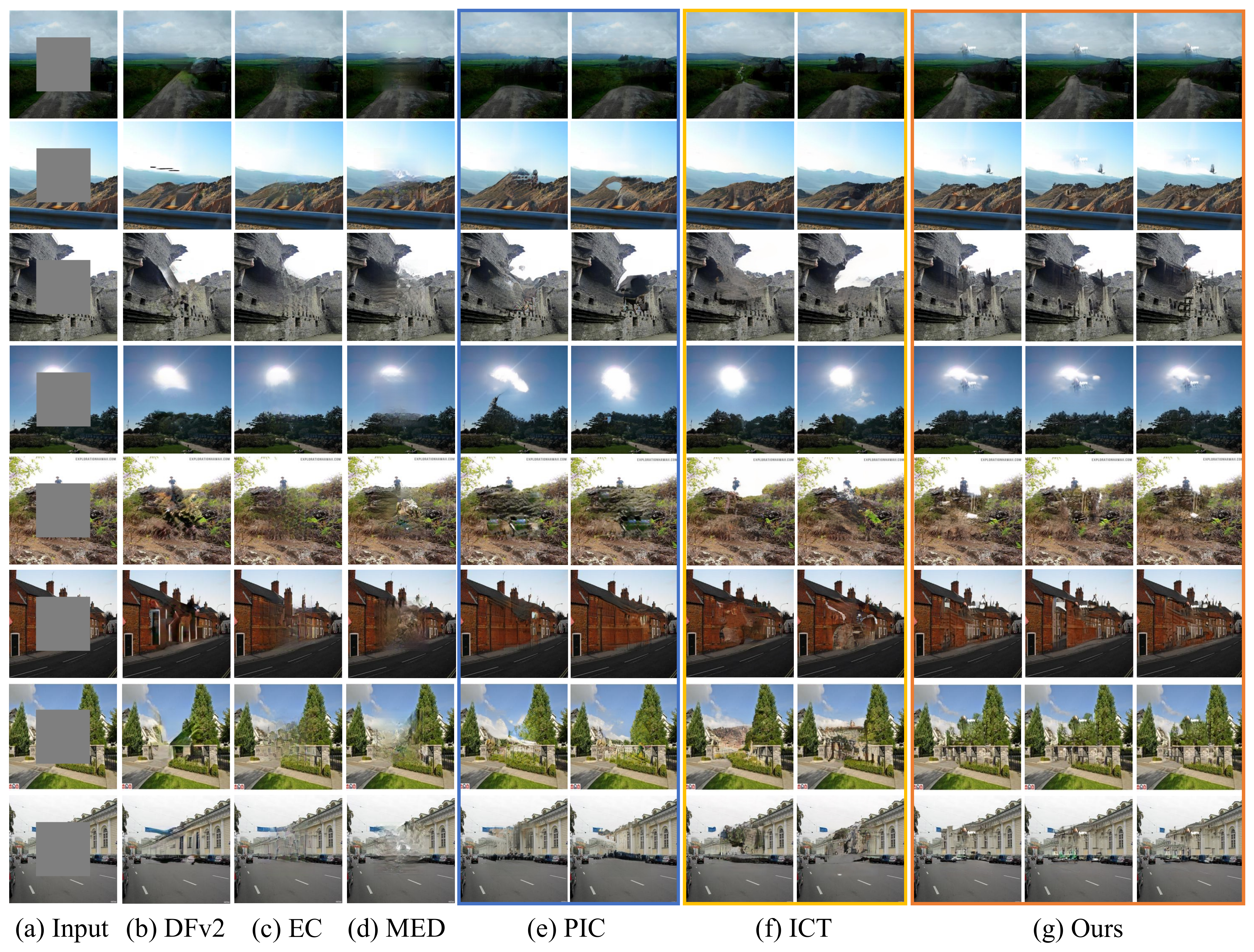}
    \caption{Pluralistic image completion results comparison with baselines. The original images come from Places2. Best viewed by zooming in.}
    \label{fig:supp_place}
\end{figure*}

\begin{figure*}[!h]
    \centering
    \includegraphics[width=0.95\textwidth]{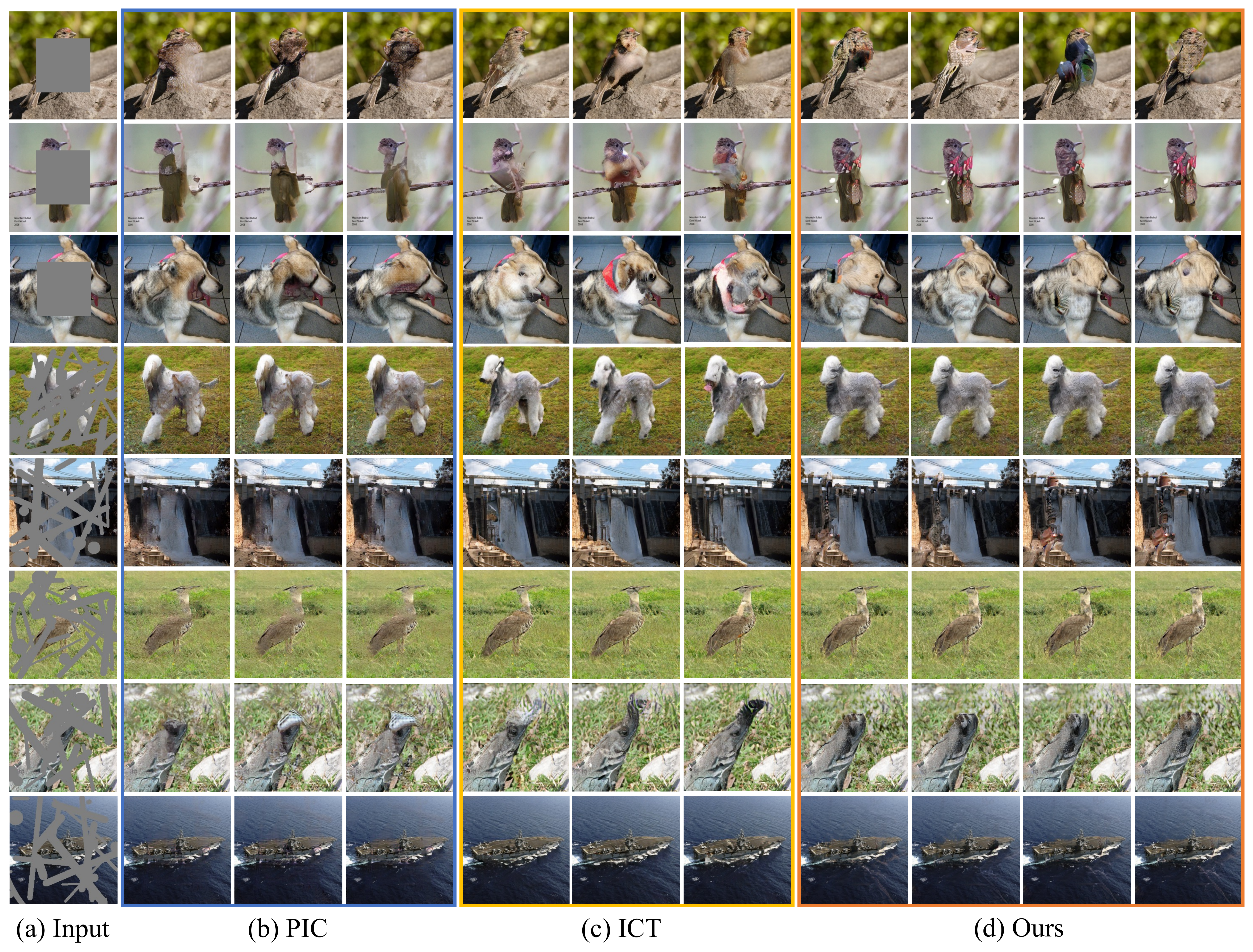}
    \caption{Pluralistic image completion results comparison with baselines. The original images come from ImageNet. Best viewed by zooming in.}
    \label{fig:supp_imagenet}
\end{figure*}

In the main paper, we provide some comparisons of pluralistic completion results with state-of-the-art methods. We provide more comparisons here. The comparison methods include DFv2\footnote{\href{https://github.com/JiahuiYu/generative\_inpainting}{https://github.com/JiahuiYu/generative\_inpainting}}, EC\footnote{\href{https://github.com/knazeri/edge-connect}{https://github.com/knazeri/edge-connect}}, MED\footnote{\href{https://github.com/KumapowerLIU/Rethinking-Inpainting-MEDFE}{https://github.com/KumapowerLIU/Rethinking-Inpainting-MEDFE}}, PIC\footnote{\href{https://github.com/lyndonzheng/Pluralistic-Inpainting}{https://github.com/lyndonzheng/Pluralistic-Inpainting}}, and ICT\footnote{\href{https://github.com/raywzy/ICT}{https://github.com/raywzy/ICT}}. The experimental results on  CelebA-HQ are provided in Figures~\ref{fig:supp_celeba_1} and \ref{fig:supp_celeba_2}. The experimental results on FFHQ, Paris StreetView, and Places2 are shown in Figures~\ref{fig:supp_ffhq}, \ref{fig:supp_paris}, and \ref{fig:supp_place} respectively. In addition, the experimental results on ImageNet are presented in Figure~\ref{fig:supp_imagenet}.

\vspace{-25pt}


\end{document}